\newif\ifshaphered

\ifdefined\isshaphered
\shapheredtrue
\fi

\documentclass[10pt,journal,compsoc]{IEEEtran}
\ifCLASSOPTIONcompsoc
  \usepackage[nocompress]{cite}
\else
  \usepackage{cite}
\fi
\ifshaphered
\usepackage{longtable}
\newcommand\revision[1]{#1}

\newcommand\revisionminor[1]{\textcolor{blue}{#1}}
\newcommand\deleted[1]{\st{#1}}

\else

\newcommand\revision[1]{#1}

\newcommand\revisionminor[1]{#1}
\newcommand\deleted[1]{}

\fi

\newcommand*{\textlabel}[2]{%
  \phantomsection
  #1\label{#2}
}
\usepackage{hyperref}

\usepackage{tikz}
\usepackage[misc]{ifsym}
\usepackage{amsmath,amssymb,amsfonts}
\usepackage{algorithmic}
\usepackage{booktabs}       
\usepackage[ruled,lined,linesnumbered]{algorithm2e}  
\usepackage{algorithmic}
\usepackage{amsthm}
\usepackage{numprint}
\usepackage{float}
\usepackage{multirow}
\usepackage{tabularx}  
\usepackage{mathrsfs}
\usepackage{balance}
\usepackage{graphicx}
\usepackage{subcaption}
\usepackage{enumerate}
\usepackage{enumitem,kantlipsum}
\usepackage{pgfplots}
\usepackage{soul}

\newtheorem{theorem}{Theorem}
\newtheorem{lemma}{Lemma}
\newtheorem{definition}{Definition}

\newcommand{\name}{kCAGC}

\newcommand{\Cli}{\mathcal{P}}

\newcommand{\cluster}{{U}}

\newcommand{\set}{\mathbb{C}}
\newcommand{\lc}{\hat{k}}

\newcommand{\x}{{\mathbf{x}}}
\newcommand{\X}{{\mathbf{X}}}

\newcommand{\mean}{{\mathbf{\mu}}}

\pgfplotsset{compat=1.18}
\begin{document}

\ifshaphered

\onecolumn
\bgroup
\def\arraystretch{1.5}

\section*{Summary of Changes}
We would like to extend our sincere gratitude to the reviewers and the editor for their detailed and insightful feedback.
We are grateful for their recognition of our efforts in improving the paper and their insightful comments which have significantly contributed to enhancing the quality of our paper.
We have endeavored to address all comments and hope that our responses meet the expectations of the reviewers and the editor. 

In the following table, we list the comments that need to be addressed. 
In the 3rd column (``Response'') we explain how we addressed each comment. The 4th column (``Link'')  is a clickable link that points to the place in the paper corresponding to the comment/response.
If a review is addressed at multiple locations in the paper, multiple links are provided. 
Each link points to the text that is added (or changed) to address the corresponding comment.
Each link is named by the initial fragment of the text it points to.
To ease the tracking of changes, we marked the updated text in \revision{blue}.




\begin{longtable}{p{.065\textwidth} | p{.3\textwidth} | p{.35\textwidth} | p{.15\textwidth}}
\hline
Reviewer & Review comment & Response & Link \\\hline

\textlabel{R-1(1)}{row:r1_1}

&

The authors have addressed the comments regarding motivations and challenges, but it is suggested to further explain the advantages of the proposed schemes (or how they address these challenges) in detail.

&

Thanks for your acknowledgments and new suggestion! 

Our advantages can be summarized as follows:

\begin{enumerate}[leftmargin=*]
\item {\bf The first privacy-preserving unsupervised method tailored for vertical settings.} As current methods usually focus on horizontal settings and supervised methods, 
we are the first to extend unsupervised graph clustering methods to vertically collaborative learning settings, which distinguishes our work from existing literature.

To achieve this design goal, we first propose a basic method that largely relies on secure aggregation to fill the gap in our setting. We then use local clusters to develop an optimized method to reduce the communication cost from $O(n)$ to $O(k^3)$, where $k$ is the number of local clusters.

\item {\bf Comparable method performance in both theory and practice.} To prove the effectiveness of our collaborative methods, we show both in theory and in practice that our methods can achieve similar performance compared to centralized methods like AGC and GCC. Furthermore, our unsupervised methods can achieve results comparable even to the semi-supervised method GraphSAGE.
\end{enumerate}

We have added a clearer description in the contribution part of the Introduction section.
&

\hyperref[MinorR1:1:1]{In horizontal settings,}

~

\hyperref[MinorR1:1:2]{However, in our vertical settings ...}

~

\hyperref[MinorR1:1:3]{To the best of  ...}

~


\\\hline

\textlabel{R-1(2)}{row:r1_2}

&

The authors have provided more experimental results. But it is suggested to explain the reason why AGC and GCC are not evaluated for L$\geq$ 4.

&
Sorry for the confusion.
AGC and GCC are centralized graph clustering methods, where $L$ is fixed at $1$. Since there is no existing unsupervised method in vertical settings, following the seminal methods~\cite{zhang2021subgraph,chen2020vertically} in supervised settings, we compare our work directly to the centralized methods.

~

In this version, we make it clear in Section~\ref{sec:experiments}. 
And we modify Table~\ref{tab:utility} to make it more clear.

&

\hyperref[MinorR1:2:1]{Since we are the first ...}

~

Table~\ref{tab:utility}



\\\hline\hline

\textlabel{R-2(1)}{row:r2_1}

&

As also noted by other reviewers, it is peculiar that an increase in 'k' leads to a decrease in accuracy rather than an improvement. This trend is counterintuitive and may suggest underlying issues with either the model's scalability or its sensitivity to parameter settings. It would be beneficial for the authors to explore and discuss why this phenomenon occurs. A deeper analysis or additional experiments might be necessary to understand the relationship between the value of 'k' and the model's performance.

&

Sorry for the confusion about the relation between $\lc$ and the accuracy. We have updated the theoretical analysis in section~\ref{sec:experiments} for better clarity. 

~

According to our theoretical analysis in section~\ref{sec:theoretical}, when $\lc = k$, our conditions are on the same scale as the centralized conditions. Therefore, when $\lc\geq k$, we can get results similar to centralized settings. 
However, the conditions we proposed are non-monotonic w.r.t $\lc$; specifically, while $||\X-\hat{C}||$ decreases, other factors increase. As a result, a larger $\lc$ does not always guarantee better results since the conditions can be more difficult to satisfy when $\lc$ increases beyond $k$.

&

\hyperref[MinorR2:1:1]{However, since ...}

\\\hline

\textlabel{R-2(2)}{row:r2_2}
&
Furthermore, the accuracy achieved by the proposed method does not exceed that of some established baselines, such as GCC. This observation is crucial, especially given that the introduction of a new method should ideally demonstrate a clear advantage over existing methods. I recommend that the authors provide a more detailed comparative analysis, possibly including a discussion on the scenarios or conditions under which their method might outperform these baselines. Additionally, insights into any potential trade-offs involved in using the proposed method compared to GCC and others would be valuable for readers and practitioners in the field.

In summary, while the manuscript has improved, addressing the issues with the results in Table 2 and providing a clearer comparative advantage over existing methods are essential steps to enhance the paper's contribution to the field.
&

GCC and AGC are both centralized graph clustering methods. 
Our method is proposed for vertically collaborative learning settings, where data is isolated by different participants. Therefore, our method is inherently less accurate compared to centralized methods.

~

Since we are the first to explore this novel yet important setting for unsupervised methods, there are no existing baselines.
Following the seminal works~\cite{zhang2021subgraph, chen2020vertically} in supervised settings, We aim to prove that our method can achieve accuracy comparable to that of centralized methods. 

~

We make it clear in Section~\ref{sec:experiments}. 
And we modify Table~\ref{tab:utility} to make the centralized baselines more clear.

&

\hyperref[MinorR1:2:1]{Since we are the first ...}

~

Table~\ref{tab:utility}

\\\hline\hline
\textlabel{R-3(1)}{row:r3_1}

&

Although this revised version has addressed my most of concerns, I still suggest it should add more statements to strengthen the relevance to the scope of this journal, especially in Section 1.

&
Thank you for your acknowledgments!
Sorry for missing the relevance.

~

We have added more statements to strengthen the relevance of TDSC. And we added more references published in TDSC. Besides collaborative learning~\cite{tdsc3TAPFed}, the security of graph data is also involved in the scope of TDSC, and we have added more related works about graph data~\cite{tdsc1Protecting, tdsc2APrivacy}.

&
\hyperref[MinorR3:1:1]{How to design ...}

~

\cite{tdsc1Protecting, tdsc2APrivacy, tdsc3TAPFed}

\\\hline
\end{longtable}

















\egroup
\twocolumn
\newpage
\else
\fi

%
\title{Attributed Graph Clustering in Collaborative\\ Settings\thanks{Research report version at~\url{https://arxiv.org/abs/2411.12329}}}
%
%
%
%

\author{Rui~Zhang,
        Xiaoyang~Hou,
        Zhihua~Tian,
        Yan~He,
        Enchao~Gong,
        Jian Liu$^{\textrm{\Letter}}$,~\IEEEmembership{Member,~IEEE,}
        Qingbiao Wu$^{\textrm{\Letter}}$,
        and~Kui Ren,~\IEEEmembership{Fellow,~IEEE}
\thanks{This work was supported by National Key Research
and Development Program of China (2023YFB2704000), National Natural Science Foundation of China (U20A20222,12271479),
and the Major Project of Science and Technology Department of Zhejiang Province (2024C01179). This work was also supported by AntGroup. (Jian Liu and Qingbiao Wu are the corresponding authors.)}
\IEEEcompsocitemizethanks{\IEEEcompsocthanksitem Rui Zhang and Qingbiao Wu are with the School 
of Mathematical Sciences, Zhejiang University, Hang Zhou, China (e-mail: {zhangrui98, qbwu}@zju.edu.cn).
\IEEEcompsocthanksitem Xiaoyang Hou, Zhihua Tian, Jian Liu and Kui Ren are with the School 
of Cyber Science and Technology, Zhejiang University, Hang Zhou, China  (e-mail: XiaoyangHou42@gmail.com, {zhihuat, liujian2411, kuiren}@zju.edu.cn).
\IEEEcompsocthanksitem Yan He and Enchao Gong are with antgroup, Hangzhou, China (e-mail: {yanhe.hy,enchao.gec}@antgroup.com).
}

}

\markboth{Journal of \LaTeX\ Class Files,~Vol.~14, No.~8, August~2015}%
{Shell \MakeLowercase{\textit{et al.}}: Bare Demo of IEEEtran.cls for Computer Society Journals}

\IEEEtitleabstractindextext{%
\begin{abstract}
Graph clustering is an unsupervised machine learning method that partitions the nodes in a graph into different groups. 
Despite achieving significant progress in exploiting both attributed and structured data information, graph clustering methods often face practical challenges related to data isolation. 
Moreover, the absence of collaborative methods for graph clustering limits their effectiveness.

In this paper, we propose a collaborative graph clustering framework for attributed graphs, supporting attributed graph clustering over vertically partitioned data with different participants holding distinct features of the same data. 
Our method leverages a novel technique that reduces the sample space, improving the efficiency of the attributed graph clustering method. 
Furthermore, we compare our method to its centralized counterpart under a proximity condition, demonstrating that the successful local results of each participant contribute to the overall success of the collaboration.

We fully implement our approach and evaluate its utility and efficiency by conducting experiments on four public datasets. The results demonstrate that our method achieves comparable accuracy levels to centralized attributed graph clustering methods. Our collaborative graph clustering framework provides an efficient and effective solution for graph clustering challenges related to data isolation.
\end{abstract}

\begin{IEEEkeywords}
attributed graph clustering, k-means, collaborative learning
\end{IEEEkeywords}
}

\maketitle


\IEEEraisesectionheading{\section{Introduction}\label{sec:introduction}}
\IEEEPARstart{A}{ttributed} graphs have been widely used in real-world applications like recommendation systems and online advertising~\cite{kolluri2021private},
\revision{\textlabel{due}{R1:2:3} to their ability for representation learning of graph-structured data such as academic networks and social media systems~\cite{sen2008collective, huang2017label, gao2018deep, fu2020magnn}.}
\revision{\textlabel{By}{R1:2:4} combining the graph structure with node information, many graph-based methods have received significant attention and achieved remarkable success in recent years~\cite{kipf2016variational, hamilton2017inductive, kipf2016semi, li2018deeper, tdsc1Protecting, tdsc2APrivacy}.}


\begin{figure}[htb]
\centering
\includegraphics[width=.95\linewidth]{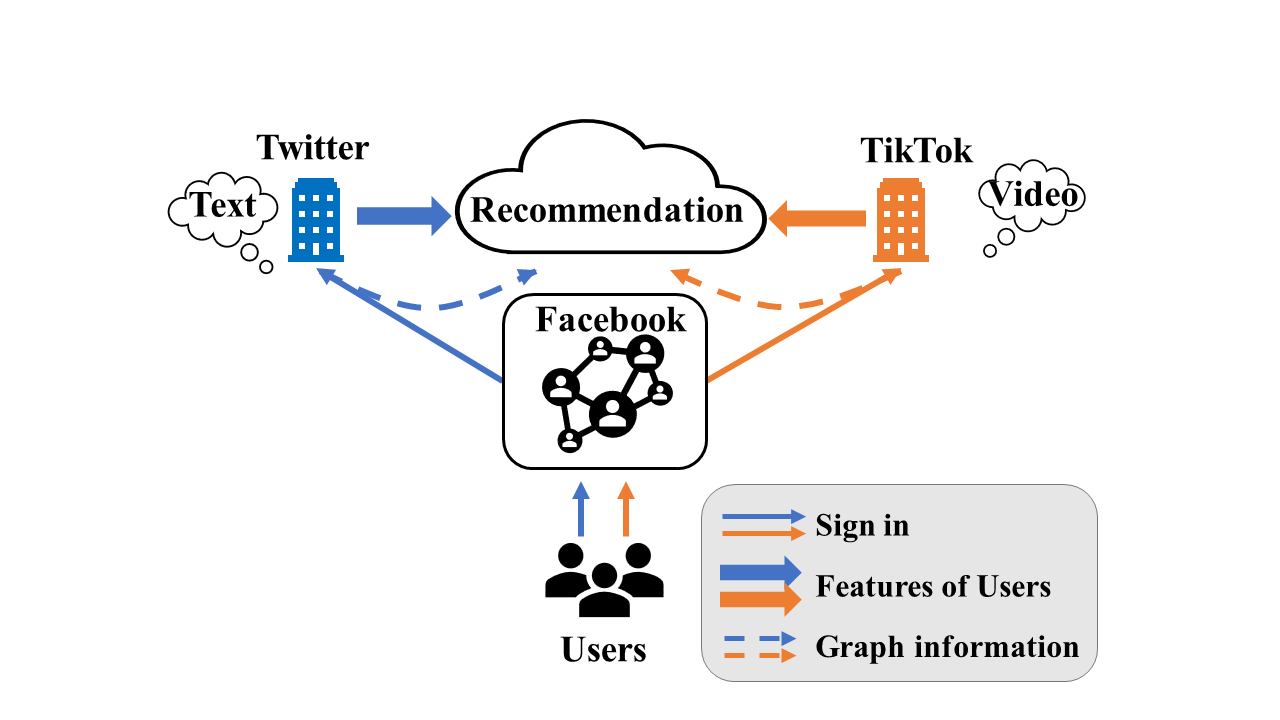}
\caption{An example of collaborative setting. Users sign in to Twitter and TikTok using their Facebook IDs. A user who wants to buy a car finds some text posts about cars on Twitter and some videos about cars on TikTok. With the shared social networks, Twitter and TikTok have access to information about cars purchased by the user's neighbors. Then they can provide better recommendations via the collaboration of text information and video information.
}
\label{fig.set}
\end{figure}

\revision{
\textlabel{However}{R1:2:1}, high-performance models usually rely on training data, which can be isolated by different participants and cannot be shared due to policy and privacy concerns~\cite {zhou2020privacy}.
Therefore, it is of great significance to discover methods that enable participants to build collaborative models while keeping the data private in each participant~\cite{wu2021fedgnn, sajadmanesh2021locally, tian2023federboost, tdsc3TAPFed}.
Collaborative learning~\cite{mcmahan2017communication} has proven to be one of the state-of-the-art solutions to the above problems. To this end, recent researchers have proposed approaches to adapt collaborative learning methods for graph data~\cite{tan2023federated, ChenLMW22, baek2023personalized}.
Most of these works focus on settings where both the graph structure and the data attributes are horizontally separated~\cite{zhou2020privacy, yao2022fedgcn}. For example, Fed-Pub~\cite{baek2023personalized} learns the similarity between participants and only updates the relevant subset of the aggregated parameters for each participant. FedSage+~\cite{zhang2021subgraph} trains a generator to replenish the missing information.
\textlabel{But}{R2:1:1} less attention has been paid to the settings where the data is vertically partitioned~\cite{mai2023vertical, qiu2023labels }.}

Consider the scenario where two companies need to collaborate to provide users with better services while keeping their data private. 
 As shown by Figure~\ref{fig.set}, apps like TikTok and Twitter will use the Facebook IDs of the users to register. Facebook is an app that users use to share and connect with their friends and people they know. In fact, Facebook users construct social networks using their Facebook IDs. With the success of Facebook, many other applications can be signed in via Facebook IDs.
Once users log into these apps with their Facebook IDs, apps like TikTok and Twitter can access the social networks constructed by Facebook and use these social networks to train a model.
\revisionminor{\textlabel{How}{MinorR3:1:1} to design reliable models that can exploit distributed data while ensuring privacy protection has raised significant concern in recent literature~\cite{tdsc3TAPFed, tian2023federboost, qiu2023labels, liu2024dynamic}.}

\revision{
\textlabel{Different}{R1:2:2} from existing studies on vertically partitioned settings~\cite{mai2023vertical, chen2020vertically, ni2021vertical}, we consider the more changeling unsupervised method where the model is exclusively trained from unlabeled data, which popular exists in practice.
Unsupervised methods on graph data, such as attributed graph clustering, have been popular and have proven to be effective in centralized frameworks
~\cite{schaeffer2007graph, wang2019attributed, cui2020adaptive}. 
\textlabel{To}{R2:1:2} the best of our knowledge, we are the first to extend the success of attributed graph clustering approaches to collaborative settings where data are vertically partitioned.}

\revision{\textlabel{A}{R1:1:1}
direct method is to employ vertically collaborative learning (federated learning). Firstly, each participant derives the local graph embedding, and then cryptographic algorithms such as secure aggregation~\cite{bonawitz2017practical} are used to concatenate the features from different participants to train the global model. In this vein, we propose a basic method that employs secure aggregation.
However, collaborative learning is known to suffer from high communication costs because of the cryptographic operations, which limit their efficiency in practice~\cite{chen2020vertically}.
\revisionminor{\textlabel{In}{MinorR1:1:1} horizontal settings, the communication cost depends on the number of participants}.
Recent works have been proposed to surpass these limitations by reducing the number of participants involved in communication~\cite{chu2022design, khan2021socially, vahidian2023efficient, liu2024dynamic}.
For instance,~\cite{chu2022design} proposes to cluster the different participants and then collaboratively updates the model parameters of the leader participant in each cluster;~\cite{khan2021socially} computes a sub-global model within each cluster, which is formed by the social relation of the participants, then they aggregate these sub-global updated parameters into the global model.} 

\revisionminor{\textlabel{However}{MinorR1:1:2}, in our vertical settings, }
the communication cost is dependent on the size of the sample space which can significantly exceed the number of participants. This is because all participants need to collaboratively compute the distance between each sample and the global model centers.~\cite{dennis2021heterogeneity} is proposed to reduce the communicating sample space of the Lloyd algorithm in horizontal settings by using local centers of participants. However, this approach directly shares the local centers, potentially leading to privacy concerns. Consequently, minimizing the communication cost in vertical settings without compromising the privacy of local data remains an open problem.

\revision{
\textlabel{To}{R1:1:3} address the aforementioned challenges, instead of reducing the number of participants in communication}, we leverage local clusters on the sample space to help identify global clusters, thus reducing the sample space and achieving accuracy comparable to centralized methods while minimizing communication costs. \revision{\textlabel{More}{R2:1:3} specifically, firstly we apply $k$-means clustering on each local participant to get the local clusters. Subsequently, we communicate the IDs of samples in each local cluster to get the intersected clusters. 
The local centers of these intersected clusters are then concatenated by secure aggregation to train the global clustering model.}
We formally prove the correctness of our algorithm compared to centralized settings using the proximity perspective~\cite{awasthi2012improved} and evaluate its performance on four public datasets, showing that it achieves comparable accuracy to centralized methods.

Our contributions can be summarized as follows:
\begin{itemize}[leftmargin=*]
    \item We introduce $\name$, a communication-efficient collaborative learning framework to solve attributed graph clustering in vertically collaborative settings where different participants sharing a common graph structure hold different features of the same data. \revisionminor{\textlabel{To}{MinorR1:1:3} the best of our knowledge, this is the first unsupervised method that addresses the attributed graphs in vertical settings.}
    \item We improve the communication efficiency by communicating only the local clusters, making the communication cost \revisionminor{reduce from $O(n)$ to $O(k^3)$, where k is the number of local clusters and $n$ is the size of dataset}.
    \item We extend the classic proximity based theoretical framework~~\cite{awasthi2012improved} for clustering to collaborative settings. Theoretically, we show that our method performs similarly to centralized methods when the local data of the participants can satisfy our proposed ``restricted proximity condition``. Specifically, the proposed ``restricted proximity condition`` is a compromise between the conditions proposed by~\cite{awasthi2012improved} and~\cite{kumar2010clustering}.
    \item  We conduct extensive experiments to verify the utility of $\name$. \revisionminor{The results show that $\name$ can achieve similar performance compared to centralized methods~\cite{zhang2019attributed, fettal2022efficient} and $\name$ is even comparable to the semi-unsupervised GraphSAGE models~\cite{hamilton2017inductive}} . 
\end{itemize}

\section{Preliminaries}\label{sec:preliminary}
\subsection{k-means clustering for attributed graphs.} 
For a given attributed graph, the structure information is fixed and independent of model training, as it relies on the adjacency matrix and its node features. Since the graph convolutions are based on the Laplacian which is computed by the adjacency matrix, the graph information filter period can be separated from the clustering period. A common practice is to perform $k$-means on the filtered graph~\cite{zhang2019attributed, zhang2021spectral}(cf. Protocol~\ref{alg:centralized} ).

Given a data matrix $\X\in \mathbb{R}^{n\times M}$ output by the graph convolution period, $k$-means can partition $n$ nodes (each node $\X_i$ has $M$ features) into $k$ clusters $\{\cluster_1, \cluster_2, \cdots, \cluster_K\}$, s.t. the following objective function is minimized:
\begin{equation}
\label{equation.1}
    \sum_{k=1}^K\sum_{i \in \cluster_k}\left\| \X_i - \mean_k \right\|^2,
\end{equation}
where $\mu_k$ is the  mean of the nodes in $\cluster_k$, 
and 
$\left\| \X_i - \mean_k \right\|^2 = \sum\limits_{j=1}^M(\X_{i, j} - \mean_{kj})^2$ 
(i.e., the square of Euclidean distance between $\x$ and $\mean_k$).

\renewcommand*{\algorithmcfname}{Protocol}
\renewcommand{\algorithmiccomment}[1]{$\triangleright$ #1}
\begin{algorithm}[!t]
\caption{Local attributed graph clustering based on $k$-means~\cite{awasthi2012improved}}
\label{alg:centralized}
\small
\KwIn {feature matrix $\X^l$}

\KwOut{$\lc$ clusters: $\{\cluster^l_1, \cluster^l_2, \cdots, \cluster^l_{\lc} \}$ and their centers $\{\mean^l_1, \mean^l_2, \cdots, \mean^l_{\lc}\}$}
Decompose 
$L_s=Q\Lambda Q^{-1}$ \\
Calculate 
$G =Q p(\Lambda)Q^{-1}$\\
Calculate 
$\X^l = G\X^l$\\
Project $\X^l$ onto the subspace spanned by the top $\lc$ singular vectors to get $\hat{\X}^l$. Run any standard 10-approximation algorithm on the projected data and estimate $\lc$ centers $(\mean^l_1, \mean^l_2, \cdots, \mean^l_{\lc})$

\For(\hfill\CommentSty{for each sample}){$i = 1 \to n$}{
    \For(\hfill\CommentSty{for each cluster}){$r = 1 \to \lc$}{
     Calculate $d_{i, r} : = ||\hat{\X}^l_i - \mean^l_r||_2^2$
    }
    Assign  $\X_i^l$ to the cluster $S_{r}$ such that  $d_{i, r}\leq \frac{1}{9}d_{i, s}$ $\forall s \neq r$
}
\For(\hfill\CommentSty{for each cluster}){$r = 1 \to \lc$}{
        Calculate $\mean^l_r := \frac{1}{|S^l_r|}\sum\limits_{i \in S_r}\X^l_i$
}
\For(\hfill \CommentSty{for each round}){$t = 1\to Q$}{
    \For(\hfill\CommentSty{for each sample}){$i = 1 \to n$}{
        \For(\hfill\CommentSty{for each cluster}){$r = 1 \to \lc$}{
            Calculate $d_{i, r} : = ||\X^l_i - \mean^l_r||_2^2$

        }
        Assign $\X^l_i$ to the cluster $\cluster^l_{r}$ such that  $d_{i, r}\leq d_{i, s}$ $\forall s \neq r$
    }
    \For(\hfill\CommentSty{for each cluster}){$r = 1 \to \lc$}{
            Calculate $\mean^l_r := \frac{1}{|\cluster^l_r|}\sum\limits_{i \in \cluster^l_r}\X^l_i$
    }
}
\end{algorithm}

Assuming there exists a target optimal partition,
Kumar and Kannan~\cite{kumar2010clustering} proposed a distribution-independent condition and proved that if their condition is satisfied, all points are classified correctly. 
Awasthi and Sheffet~\cite{awasthi2012improved} improved this work by introducing a weaker condition with better error bounding. 
More recently, $k$-FED\cite{dennis2021heterogeneity} successfully extended these results to horizontal collaborative settings, where participants’ data share the same feature space but differ in nodes. $k$-FED proposed new conditions suitable for the horizontal collaborative setting, although stricter than those in~\cite{awasthi2012improved}.
Inspired by their work, we provide a correctness analysis based on a restricted proximity condition for collaborative settings, where participants have different features of the same data.



\subsection{Secure aggregation.}
Bonawitz, et al.~\cite{bonawitz2017practical} propose a secure aggregation protocol to protect the local gradients in Google's horizontal FL.
\revision{
This protocol, a cryptographic algorithm, is designed to compute the sums of vectors without disclosing any information other than the aggregated output.
}
They use threshold secret sharing to handle dropped-out participants.
Instead, we assume the participants are large organizations that are motivated to build the clusters, so they will not drop out in the middle of the protocol. 
Based on this assumption, we significantly simplify the secure aggregation protocol (cf. Protocol~\ref{alg:aggregation} \revision{in Appendix~\ref{appdix.B}\footnote{The appendix is in our report version.}}).

Specifically, $\Cli_l$ initializes a large prime $p$ and a group generator $g$ for Diffie-Hellman key generation.
It also initializes a small modular $N$ for aggregation: the aggregation works in $\mathbb{Z}_N$.
However, the gradients are floating-point numbers. 
To deal with this, we scale the floating-point numbers up to integers by multiplying the same constant to all values and dropping the fractional parts. 
This idea is widely used in neural network training and inferences~\cite{secureML, minionn}.
$N$ must be large enough so that the absolute value of the final sum is smaller than $\left\lfloor N/2 \right\rfloor$.

\section{The Proposed Methods}
\label{sec:collaborative}
\subsection{Problem Statement}

\subsubsection{Centralized Settings}
Given an attributed non-directed graph $\mathcal{G=(V, E, \X)}$, where $\mathcal{V}=\{v_1, v_2, \dots, v_n\}$ is the vertex set with $n$ nodes in total, $\mathcal{E}$ is the edge set, and $\X\in\mathbb{R}^{n\times m}$ is the feature matrix of all nodes.

We aim to partition the nodes in $\X$ in different clusters $\mathcal{U}=\{U_1, U_2, \cdots, U_k\}$.

\subsubsection{Collaborative settings}
In collaborative settings, the feature matrix $\X$ is divided into $L$ parts $\X=[\X^1, \cdots, \X^L]$. Recall the scenario mentioned in {\em {Introduction}}, each participant $\Cli^l$ holds feature sets $\X^l$ respectively and share same graph $\mathcal{G=(V, E, \cdot)}$. Same with the centralized settings, we aim to partition the nodes in $\X$ into different clusters $\mathcal{U}=\{U_1, U_2, \cdots, U_k\}$.
{\bf We assume that the record linkage procedure has been done already}, i.e., all $L$ participants know that their commonly held samples are $\x_{1}\cdots\x_{n}$.
We remark that this procedure can be done privately via {\em multi-party private set intersection}~\cite{10.1145/3133956.3134065}, which is orthogonal to our paper.

We assume there is a secure channel between any two participants, hence it is private against outsiders.
The participants are incentivized to build the clusters (they will {\em not} drop out in the middle of the protocol), but they want to snoop on others' data.
We assume a threshold $\tau < L - 1$ number of participants are compromised and they can collude with each \revision{\textlabel{other}{R2:5:1}~\cite{chen2020vertically, tian2023federboost}.}

Our key idea to deal with the privacy preserving challenge is twofold. First, we propose a method that make use of the local information to reduce the scale of communication and then we use secure aggregation to protect privacy for data during communication.

\subsection{Attributed Graph}\label{sec.attributed_graph}

In this section, we introduce our method to extract the graph structure information. We assume that nearby nodes on the graph are similar to each other.
We analyze the intuition of designing the graph filter from the perspective of the Fourier transform. This perspective, which relates the above assumption and the design of the graph filter, is more fundamental and intuitive compared to the approaches used~\cite{zhang2019attributed} and~\cite{cui2020adaptive} in which they use Laplacian-Beltrami operator and Rayleigh quotient respectively.
We aim to design a graph filter $G$ so that nearby nodes have similar node embeddings $G\X$.
We then explore the concept of smoothness from the perspective of the Fourier transform.

Given an attributed non-directed graph $\mathcal{G=(V, E, \X)}$, the normalized graph Laplacian is defined as 
\begin{equation}
    L_s = I-D^{-\frac{1}{2}}AD^{-\frac{1}{2}},
\end{equation}
where $A$ is an adjacency matrix and $D$ is a diagonal matrix called the degree matrix. 
Since the normalized graph Laplacian $L_s$ is a real symmetric matrix,
$L_s$ can be eigen-decomposed as $L_s=Q\Lambda Q^{-1}$, where $\Lambda =diag(\lambda_1, \dots, \lambda_n)$ is a diagonal matrix composed of the eigenvalues, and $Q = [q_1, \dots, q_n]$ is an orthogonal matrix composed of the associated orthogonal eigenvectors.

The smoothness of a graph signal $f$ can be deduced similar to the classic Fourier transform.
The classical Fourier transform of a function $f$ is,
\begin{equation}
    \hat{f}(\xi) = \int_{\mathbb{R}}f(t)e^{-2\pi i \xi t}dt,
\end{equation}
which are the eigenfunctions of the one-dimensional Laplace operator,
\begin{equation}
    -\Delta(e^{-2\pi i \xi t}) = -\frac{\partial^2}{\partial t^2}e^{-2\pi i \xi t} = (2\pi i \xi)^2e^{-2\pi i \xi t}.
\end{equation}
For $\xi$ close to zero, the corresponding eigenfunctions are smooth, slowly oscillating functions, whereas for $\xi$ far from zero, the corresponding eigenfunctions oscillate more frequently. Analogously, the graph Fourier transform of function $f$ on the vertices of $\mathcal{G}$ is defined as,
\begin{equation}
    \hat{f}(\lambda_l) =\sum_{i=1}^nf(i)q_l(i). 
\end{equation}
The inverse graph Fourier transform is defined as,
\begin{equation}
    f(i) :=\sum_{i=1}^{n-1}\hat{f}(\lambda_l)q_l(i). 
\end{equation}
Smaller values of $\lambda_l$ will make the eigenvectors vary slowly across the graph, meaning that if two nodes are similar to each other, the values of the eigenvectors at those locations are more likely to be similar. If the eigenvalue $\lambda_l$ is large, the values of the eigenvectors at the locations of similar nodes are likely to be dissimilar. Based on the assumption that nearby nodes are similar, we need to design a graph filter with small eigenvalues to replace the Laplacian.

A linear graph filter can be represented as
\begin{equation}
    G =Q p(\Lambda)Q^{-1}\in R^{n\times n},
\end{equation}
where $p(\Lambda) = diag(p(\lambda_1), \dots, p(\lambda_n))$. 
According to the above analysis, $p(\cdot)$ should be a nonnegative decreasing function. Moreover,~\cite{chung1997spectral} proves that $\lambda_l \in [0,2]$. Then $p(\cdot)$ becomes easier to design.

Recall that the Laplacian is fixed for all participants since they share the common graph structure  $\mathcal{G=(V, E, \cdot)}$. We can design the same graph filter $G\in \mathbb{R}^{n\times n}$ for all participants. 
We use an adaptive graph filter $p(\lambda)=(1-\frac{\lambda}{2})^\psi$~\cite{zhang2019attributed} and $p(\lambda)=(1-\frac{\lambda}{||L_s||})^\psi$ for our experiments, where $\psi$ is a hyperparameter. Since we use the graph filter to get the node embeddings, we refer to the resulting embeddings as filtered feature sets for consistency.

For participant $\Cli_l$, it holds feature set $\X^l \in \mathbb{R}^{n\times m_{l}}$. Similar to the centralized settings, applying graph convolution on the feature set $\X^l$ locally on participant $\Cli_l$ yields the filtered feature set (node embeddings),
\begin{equation}
    \overline{\X^l}=G\X^l,
\end{equation}
where $l=1,2, \dots, L$. After each participant performs the graph filter on their local data, we obtain the filtered feature sets $\overline{\X}^1, \cdots, \overline{\X}^L$. We then use the filtered feature sets for our clustering method.
For the sake of clarity, we overload the symbols $\X^1, \cdots, \X^L$ to denote $\overline{\X}^1, \cdots, \overline{\X}^L$ for the rest of the paper.
With these filtered data $\X$, we transform the graph clustering problem into a traditional clustering problem.

\subsection{Basic method.}

In collaborative settings, since
all participants have the common graph structure information, they can perform the same period of graph information filtering ($G$) to their own data ($\X^l$).

However, in order to cluster these filtered nodes into different clusters, 
participants have to {\em jointly} run the Lloyd step as the filtered graph features are distributed. 
Specifically, each $\Cli$ has all nodes to {\em locally} update the centers corresponding to its features. They need to cooperate with other participants for every node when computing the distance between the node and each cluster as mentioned in (~\ref{equation.1}). After finding the nearest cluster for all the nodes, all participants can update the cluster centers locally.

Naively, we can still have all participants jointly compute Euclidean distances between each node and each centroid 
using secure aggregation. 

\revision{
\textlabel{In}{R4:3:1} the centralized Protocol~\ref{alg:centralized}, the protocol initiates with applying the graph filter described in Section~\ref{sec.attributed_graph} to the features $\X$. Subsequently, Singular Value Decomposition (SVD) is employed to project $\X$ onto the top $k$ subspace. Following this, a 10-approximation algorithm determines the initial centers. The distance between node $i$ and center $r$ is then computed as $d_{i,r}=||\X_i-\mean_r||^2$. Subsequently, node $i$ is assigned to cluster $r$ if $d_{i, r}\leq \frac{1}{9}d_{i, s}, \forall s\neq r$. And the centers are updated as the average of nodes within this cluster. Finally, Lloyd's algorithm is applied to get the output clusters until reaching the maximum number of iterations or convergence,
\begin{itemize}
    \item $\cluster_r \leftarrow \{i:d_{i,r}\leq d_{i,s}, \forall s\neq r\}$
    \item $\mean_r \leftarrow \mean(\cluster_r)$
\end{itemize}
}

\revision{\textlabel{In}{R4:3:3} basic \name, it is essentially the same with Protocol~\ref{alg:centralized}. We first apply the graph filter, SVD, and the 10-approximation algorithm to each participant's local feature. For the following part, denote $\hat{\sum}$ as the secure aggregation  we can simply replace the operation between different participants with secure aggregation, $
    d_{i,r}\rightarrow \hat{\sum}_{l=1, \cdots,L}(d_{i_r}^l)$. 
For the first step of Lloyd's algorithm, the assignment is shared between participants directly because it is the target output of $\name$. For the second step, the update of centers can be conducted locally for each participant.
}

Protocol~\ref{alg:collaborative_basic} depicts this solution. 
This basic solution requires $Kn$ secure aggregations for each iteration, which is expensive given that the number of nodes $n$ is usually large (each organization might have local data that is in Petabytes).

\renewcommand*{\algorithmcfname}{Protocol}
\renewcommand{\algorithmiccomment}[1]{$\triangleright$ #1}
\begin{algorithm}[!t]
\caption{Basic collaborative attributed graph clustering based on $k$-means (basic $\name$)}
\label{alg:collaborative_basic}
\small
\KwIn {each $\Cli_l$ inputs  feature $\X^l$, convolution order $t$ 

\CommentSty{for simplicity, we assume each $\Cli_l$ holds a single feature, 
hence $m=L$}}
\KwOut{$k$ clusters: $\{\cluster_1, \cluster_2, \cdots, \cluster_k\}$ and their centers $\{\mean_1, \mean_2, \cdots, \mean_k\}$} 

\For(\hfill\CommentSty{for each $\Cli$}){$l=1\to L$}{
    Decompose 
    $L_s=Q\Lambda Q^{-1}$ \\
    Calculate 
    $G =Q p(\Lambda)Q^{-1}$\\
    Calculate 
    $\X^l = G\X^l$\\
            Project $\X^l$ onto the subspace spanned by the top $\lc$ singular vectors to get $\hat{\X}^l$. \\
            Run any standard 10-approximation algorithm on the projected data and estimate $k$ centers $(\mean^l_1, \mean^l_2, \cdots, \mean^l_{\lc})$
}

\For(\hfill\CommentSty{for each node}){$i = 1 \to n$}{
        \For(\hfill\CommentSty{for each cluster}){$r = 1 \to k$}{
            \For(\hfill\CommentSty{for each $\Cli$/feature}){$j = 1 \to m$}{
                $\Cli_j$ calculates $d^j_{i, r} : = (\hat{\X}_{i}^j - \mean_{r}^j)^2$
            }
            all $\Cli$s run {\bf secure aggregation}: 
            $D_{i, r}:=\sum\limits_{j=1}^Md^j_{i, r}$
        }
        $\Cli_L$ assigns $\X_i$ to the cluster $S_{q}$ such that  $D_{i, q}\leq \frac{1}{9}D_{i, p}$ $\forall p \neq q$; and tells other $\Cli$s the assignment
}
\For(\hfill\CommentSty{for each cluster}){$r = 1 \to k$}{
        \For(\hfill\CommentSty{for each $\Cli$/feature}){$j = 1 \to m$}{
            $\Cli_j$ calculates $\mean_{k}^j := \frac{1}{|S_k|}\sum\limits_{i \in S_k}\X^j_i$
        }
}

\For(\hfill \CommentSty{for each round}){$t = 1\to Q$}{
    \For(\hfill\CommentSty{for each node}){$i = 1 \to n$}{
        \For(\hfill\CommentSty{for each cluster}){$r = 1 \to k$}{
            \For(\hfill\CommentSty{for each $\Cli$/feature}){$j = 1 \to m$}{
                $\Cli_j$ calculates $d^m_{i, r} : = (\X^j_i - \mean^j_r)^2$
            }
            all $\Cli$s run {\bf secure aggregation}: 
            $D_{i, r}:=\sum\limits_{j=1}^Md^j_{i, r}$
        }
        $\Cli_L$ assigns $\X_i$ to the cluster $\cluster_{q}$ such that  $D_{i, q}\leq D_{i, p}$ $\forall p \neq q$; and tells other $\Cli$s the assignment
    }
    \For(\hfill\CommentSty{for each cluster}){$r = 1 \to k$}{
        \For(\hfill\CommentSty{for each $\Cli$/feature}){$j = 1 \to m$}{
            $\Cli_j$ calculates $\mean_r^j := \frac{1}{|\cluster_r|}\sum\limits_{i \in \cluster_r}\X_i^j$
        }
    }
}
\end{algorithm}

\subsection{Optimized method.}\label{subsec.optimized_method}

As mentioned above, the basic method needs too many expensive cryptographic operations. In the optimized method, our goal is to make the number of secure aggregations independent of the number of nodes $n$. 
We do this by utilizing the local clustering results to generate a small set of nodes to reduce the sample space for communication. We provide a mathematical intuition to explain the relation between the local clustering results and the centralized results.

\subsubsection{Mathematics Intuition}

Our intuition is based on the Hyperplane separation theorem~\cite{hastie2009elements}, which states that if there are two disjoint nonempty convex subsets of $\mathbb{R}^m$, then there exists a hyperplane to separate them. The aim of our clustering method is to find the hyperplanes that can separate the nodes in a manner similar to the centralized method.

\begin{figure}[ht]
    \centering
    \begin{subfigure}{0.15\textwidth}
        \includegraphics[width=1\linewidth]{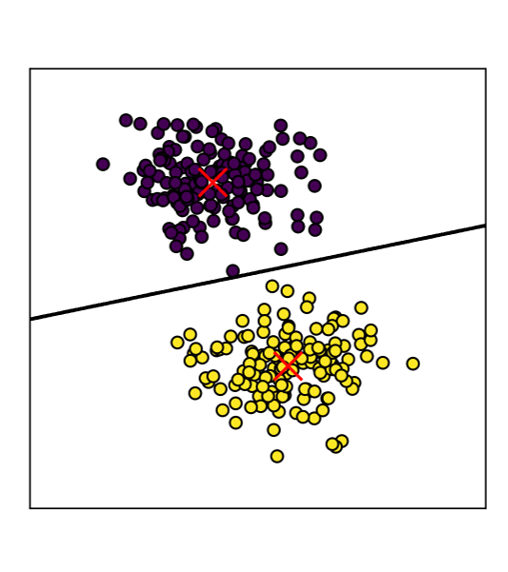}
        \caption{}
        \label{fig:cora_labels_init}
    \end{subfigure}
    \hfill 
    \begin{subfigure}{0.15\textwidth}
        \includegraphics[width=1\linewidth]{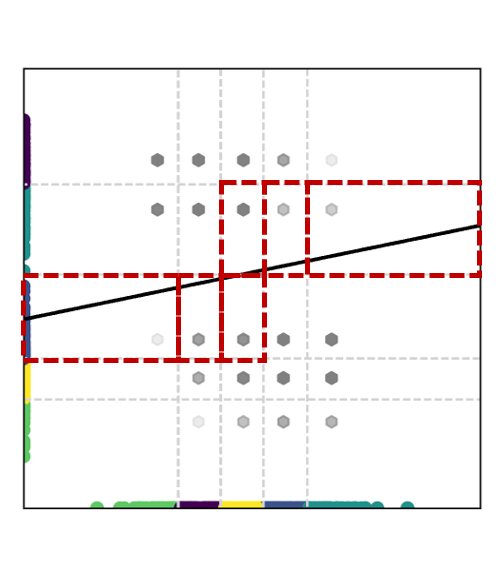}
        \caption{}
        \label{fig:cora_agc_init}
    \end{subfigure}
    \hfill 
    \begin{subfigure}{0.15\textwidth}
        \includegraphics[width=1\linewidth]{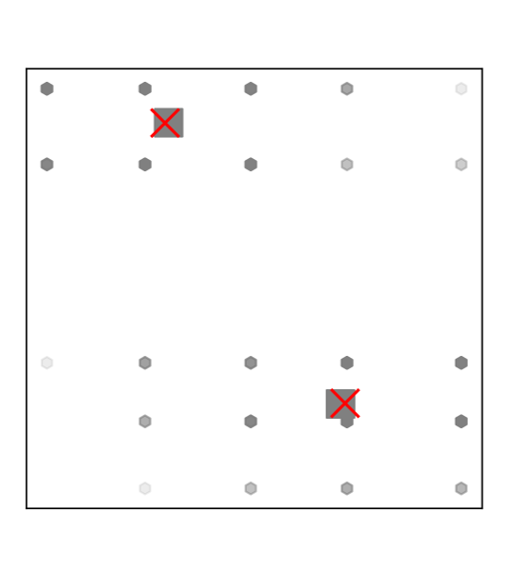}
        \caption{}
        \label{fig:cora_protocol3_init}
    \end{subfigure}
    \hfill 
    \caption{\revision{The visualization of the intuition. (a): $\mathbb{R}^2$ space separated by 1-d hyperplane. The red $\times$ are the cluster centers. (b): Local nodes are on the axes. $\mathbb{R}^2$ space is separated into $5^2$ subspaces for 5 local clusters in each dimension. Grey nodes are the centers of the subspace, darker grey nodes mean there are more nodes in this subspace. (c): This figure shows the comparisons between cluster results of the subspace (grey $\square$) and the centralized results (red $\times$)}}
        \label{fig:initui}
\end{figure}

Consider the classification problem with two classes, in the centralized method, the output clusters of Protocol~\ref{alg:centralized} on the whole space $\mathbb{R}^m$ are nonempty convex sets that can be separated by a $(m-1)-$dimension hyperplane $H$ \revision{(Figure~\ref{fig:initui}(a))}. 
\revision{\textlabel{Assuming}{R2:7:2} that each participant holds one dimension of the features,} 
if Protocol~\ref{alg:centralized} is applied to separate \textbf{each dimension} of the node features into two classes, there will be $m$ hyperplanes to separate  $\mathbb{R}^m$. In specific, the output clusters on the first dimension of node features can be separated by a hyperplane denoted as the first hyperplane, the output clusters on the second dimension can be separated by the second hyperplane, and so on.

It can be shown that $m$-dimension space can be separated into at most $2^m$ subspaces by $m$ hyperplanes~\cite{qiuchengtong}. 
If all nodes in a subspace are classified as a whole, then nodes that should be in different classes will be misclassified if they are in the same subspace. 
For the case where only part of the subspaces intersects $H$, only those subspaces intersecting with $H$ will have misclassified nodes \revision{(the red squares in Figure~\ref{fig:initui}(b)).}
Therefore, when there are enough subspaces, the proportion of subspaces intersecting with H is small, and the proportion of misclassified nodes is small.
\revision{
\textlabel{Figure}{R1:4.3:1}
~\ref{fig:initui} visualize the intuition with a tiny example in  $\mathbb{R}^2$.}

\subsubsection{k-means based Collaborative Attributed Graph Clustering}\label{subsec.kcagc}
Building upon the mathematics intuition,
our approach combines the graph structure with local data the same as the basic method.
Then each participant applies Protocol~\ref{alg:centralized} to its own feature, assigning its nodes to $\lc$ local clusters.
Subsequently, participants exchange node IDs within each cluster and compute intersections to construct $\lc^L$ clusters ($L$ is the number of participants). For each cluster, a virtual node is constructed locally by the center of the \revision{intersected} cluster. Finally, Protocol~\ref{alg:collaborative_basic} after line~6 is applied to these $\lc^L$ nodes by all participants. \revision{In specific, \textlabel{security}{R4:3:2} aggregation is applied to compute the concatenate distance between each virtual node and each center. Then each virtual node is classified into the nearest clusters and $\Cli_L$ shares the IDs of each cluster. The centers of the clusters are updated by computing the average of virtual nodes in the corresponding cluster locally. After the iteration is completed, the nodes in each intersected cluster are classified into clusters of the corresponding virtual node.
}
This reduces the complexity of communication from $n$ to $\lc^L$. 

\renewcommand*{\algorithmcfname}{Protocol}
\renewcommand{\algorithmiccomment}[1]{$\triangleright$ #1}
\begin{algorithm}[!t]
\caption{Optimized collaborative attributed graph clustering based on $k$-means (k-CAGC)}
\label{alg:collaborative_optimized}
\small
\KwIn {each $\Cli_l$ inputs feature $\X^l $ 

\CommentSty{for simplicity, we assume each $\Cli_l$ holds a single feature, 
hence $m=L$}}
\KwOut{$k$ clusters: $\{\cluster_1, \cluster_2, \cdots, \cluster_k\}$ and their centers $\{\mean_1, \mean_2, \cdots, \mean_k\}$}

\For(\hfill\CommentSty{for each $\Cli$}){$l = 1 \to L$}{
    $\Cli_l$ locally runs Protocol~\ref{alg:centralized} on $\X^l$ to partition it into $\lc$ clusters: $\{\hat{\cluster}_1^l, \cdots, \hat{\cluster}_{\lc}^l\}$; sends node {\bf ID}s in each cluster to $\Cli_L$
}

$\Cli_L$ sets $\set := \{\hat{\cluster}_1^1, \cdots, \hat{\cluster}_{\lc}^1\}$

\For(\hfill\CommentSty{for each feature}){$l = 2 \to L$}{
    $\Cli_L$ initializes $\set'$ as empty
    
    for each cluster $\hat{\cluster}$ in $\set$ and each cluster $\hat{\cluster}'$ in $\{\hat{\cluster}_1^l, \cdots, \hat{\cluster}_{\lc}^l\}$, $\Cli_l$ adds $\hat{\cluster} \cap \hat{\cluster}'$ to $\set'$
    
    $\Cli_L$ sets $\set := \set'$
}
$\Cli_L$ sends $\set = \{{\cluster}_1, \cdots, {\cluster}_{\lc^L}\}$ to other $\Cli$s

\For(\hfill\CommentSty{for each $\Cli$}){$l = 1 \to L$}{
    \For(\hfill\CommentSty{for $\hat{\cluster}$ in $\set$}){$i = 1 \to \lc^L$}{
                $\Cli_l$ set $y^l_i$ as the corresponding $\mean(\hat{\cluster}_j^l)$, if ${\cluster}_i^l \subset \hat{\cluster}_j^l$
    }
}

All $\Cli$s run {\bf Protocol~\ref{alg:collaborative_basic}} after line~6 to get $k$ clusters $\{\cluster_1, \cdots, \cluster_{k}\}$: each $\Cli_l$'s input is $\mathbf{Y}^l = [y^l_{1}, \cdots, y^l_{\lc^L}]^T$ and
$\{|{\cluster}_1|, \cdots, |{\cluster}_{\lc^L}|\}$ 
are the weights.

Each set in $\{{\cluster}_1, \cdots, {\cluster}_{\lc^L}\}$ is assigned to a set in $\{\cluster_1, \cdots, \cluster_{k}\}$ based on its centers.   
\end{algorithm}

Protocol~\ref{alg:collaborative_optimized} shows the optimized collaborative method which we call $\name$.
We remark that cluster sizes $\{|\hat{\cluster}_1|, \cdots, |\hat{\cluster}_{\lc^L}|\}$ are used as  weights for running Protocol~\ref{alg:collaborative_basic}:
in line~18 and line~33 of Protocol~\ref{alg:collaborative_basic} (which is transformed from line~12 and line~22 of Protocol~\ref{alg:centralized}), $|S_k|$ (or $|\cluster_k|$) is calculated by sum the weights $\hat{\cluster}$ of data points in $S_k$(or $\cluster_k$), and $\X^m_i$ is replaced with $|\hat{\cluster_i}|\X^m_i$;
in line~16 of Protocol~\ref{alg:collaborative_optimized}, $|\hat{\cluster}_k|$ is calculated as all nodes in $\hat{\cluster}_k$.

\revision{
\textlabel{During}{R1:3.1:1} the communication process, secure aggregations incur significantly higher costs compared to the one-time communication of node IDs, therefore, we analyze the number of secure aggregations required for basic $\name$ and $\name$ to show the efficiency improvement achieved by our optimized method.}
The number of secure aggregations in Protocol~\ref{alg:collaborative_optimized} is $O((Q+1))k\cdot \lc^L)$.
This is suitable for scenarios where the number of participants $L$ is small.
We further optimize our method for the cases where $L$ is large.
The idea is to arrange participants into the leaves of a binary tree and run Protocol~\ref{alg:collaborative_optimized} for each internal node of the tree:
\begin{itemize}[leftmargin=*]
    \item In the first layer of the internal nodes, each pair of the participants runs Protocol~\ref{alg:collaborative_optimized} to jointly partition the nodes into $\lc$ clusters based on their features.
    Notice that the output of Protocol~\ref{alg:collaborative_optimized} here is $\{\hat{\cluster}_1, \hat{\cluster}_2, \cdots, \hat{\cluster}_{\lc}\}$.
    \item For each node in the next layer of the tree, the corresponding leaves (i.e., participants) run Protocol~\ref{alg:collaborative_optimized}  but omit lines 1-3 this time: directly calculating the intersection of two clusters output by its child nodes.
    \item They run Step 2 until reaching the root of the tree, the output of which is the final $k$ clusters.
\end{itemize}
For each internal node of the tree, they need to run Protocol~\ref{alg:collaborative_optimized}, which requires $O((Q+1)\lc^3)$ secure aggregations.
Given that the number of nodes is equal to the number of leaves (i.e., participants) minus one, the total number of secure aggregations is $O((Q+1)L\cdot \hat{k}^3)$.

\revision{
\textlabel{Recall}{R1:3.1:3} that for basic $\name$ (Protocol~\ref{alg:collaborative_basic}), it needs $nk$ secure aggregations per iteration. Assuming that basic $\name$ and $\name$ need $Q$ iterations to achieve convergence, the total number of secure aggregations for basic $\name$ amounts to $(Q+1)nk$. 
When $L=2$, assuming that each participant possesses $\lc=k$ local clusters, there will be at most $k^2$ intersections between these two participants. Consequently, the number of secure aggregations required by $\name$ would be $\frac{k^2}{n}$ times that of basic $\name$. For instance, for the ``Cora'' dataset and the ``Pubmed'' dataset, $\name$ needs only $1.8\%$ and $0.5\%$ of the secure aggregations compared to basic $\name$.
For $L=3$, the intersections increase to $k^3$ for $\name$. Then $\name$ will need $12.4\%$ and $1.4\%$ of the secure aggregations required by basic $\name$ for the ``Cora'' and the ``Pubmed'' dataset. And for $L>3$, the proportion will be $\frac{L\cdot k^3}{n}$, which remains less than $64\%$ when $L\leq5$ for ``Cora'' dataset and less than $5\%$ when $L\leq 16$ for the ``Pubmed'' dataset.}

\revision{
\textlabel{Note}{R2:11:1} that in vertical settings, the number of participants is usually less than five~\cite{yang2019federated, wei2022vertical}, we provide experiments where $L<16$ to emphasize our efficiency improvement for datasets with a larger sample space.  
}



After the training period of $\name$,
to classify a new node into a cluster in $\{\cluster_1, \cluster_2, \cdots, \cluster_k\}$, all participants need to compute the nearest cluster $\cluster_i$ to this node collaboratively. Therefore, the prediction period of $k-CAGC$ for each node needs $k$ secure aggregation.

\subsection{Theoretical analysis}
\label{sec:theoretical}

\begin{figure*}[ht]
    \centering
    \begin{subfigure}{0.24\textwidth}
        \includegraphics[width=1\linewidth]{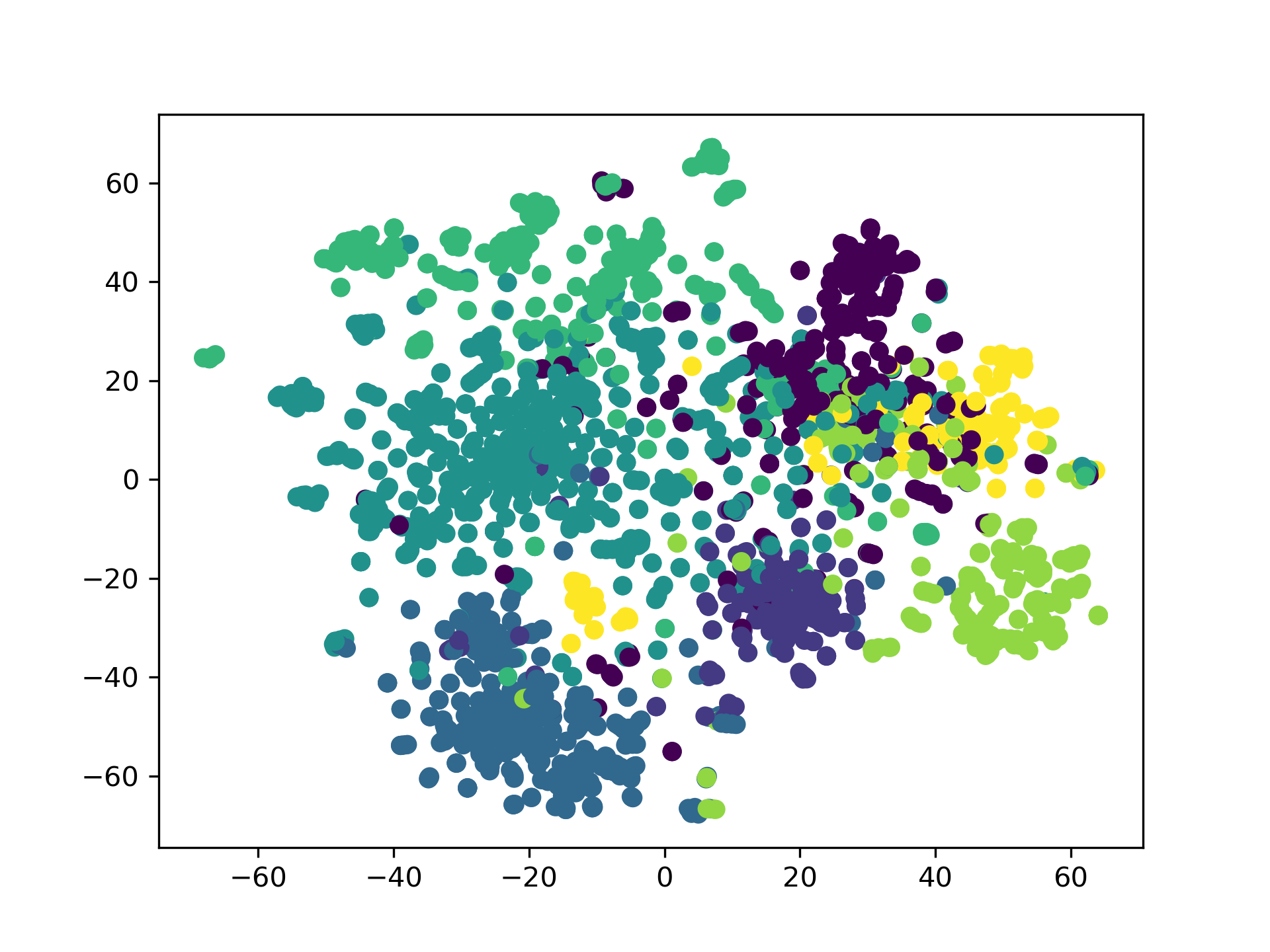}
        \caption{Real Labels}
        \label{fig:cora_labels}
    \end{subfigure}
    \hfill 
    \begin{subfigure}{0.24\textwidth}
        \includegraphics[width=1\linewidth]{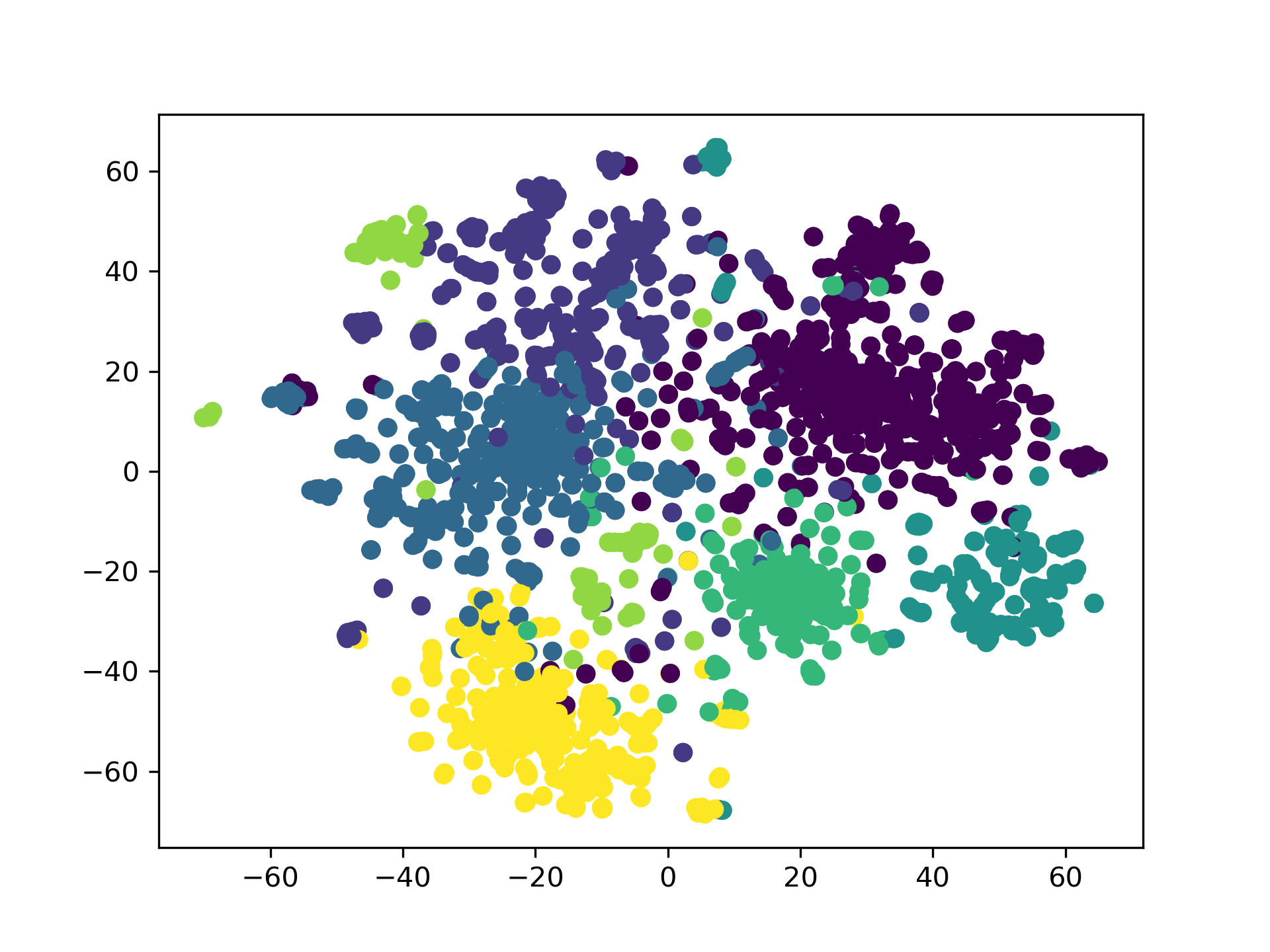}
        \caption{AGC}
        \label{fig:cora_agc}
    \end{subfigure}
    \hfill 
    \begin{subfigure}{0.24\textwidth}
        \includegraphics[width=1\linewidth]{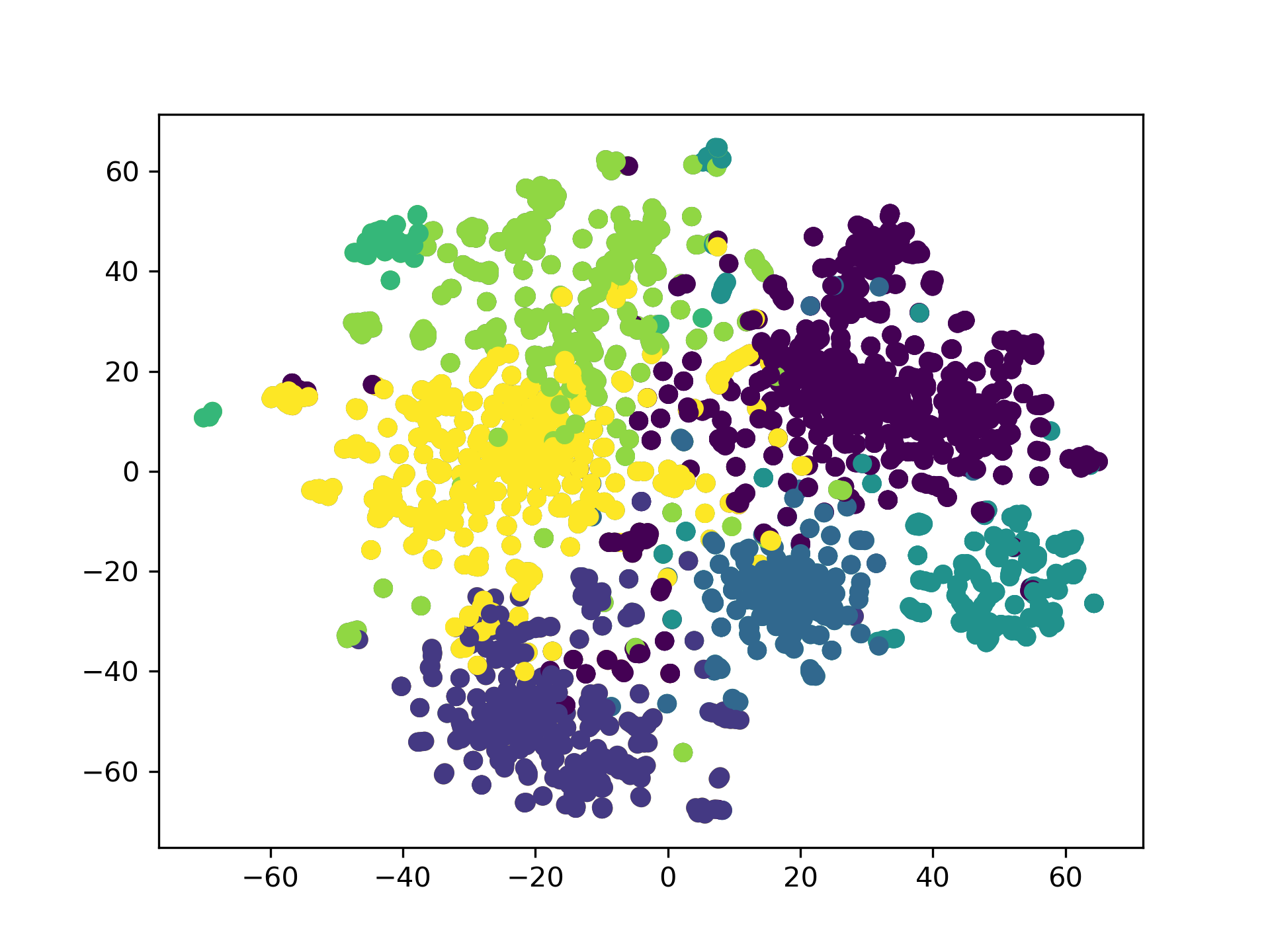}
        \caption{Protocol~\ref{alg:collaborative_basic}}
        \label{fig:cora_protocol1}
    \end{subfigure}
    \hfill 
    \begin{subfigure}{0.24\textwidth}
        \includegraphics[width=1\linewidth]{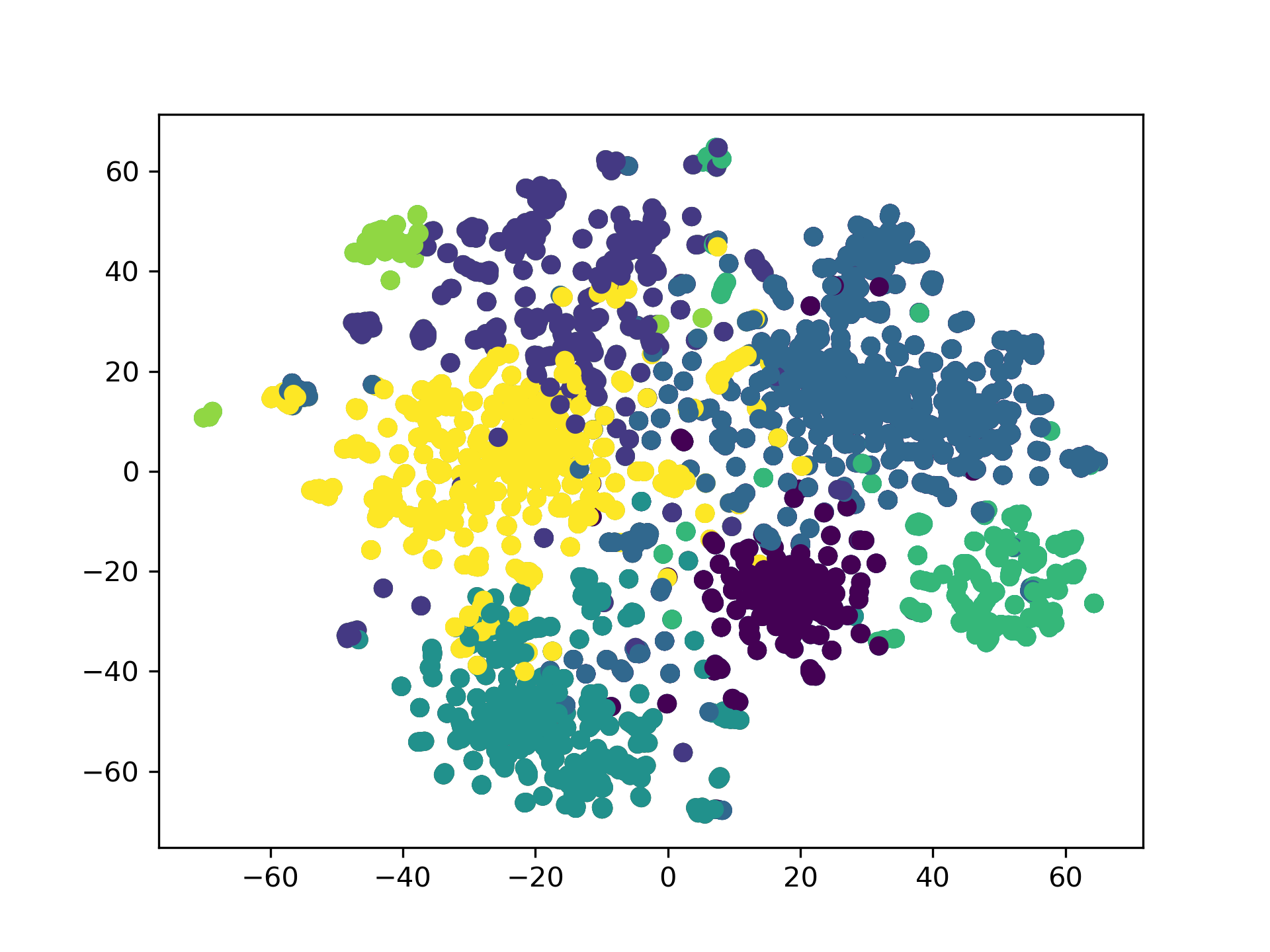}
        \caption{kCAGC}
        \label{fig:cora_kcagc}
    \end{subfigure}
\caption{The t-SNE projection of the feature for ``Cora'' dataset according to the real (a) labels, (b) AGC, (c) Protocol~\ref{alg:collaborative_basic}, and (d) $\name$. We choose L=2 for protocol\ref{alg:collaborative_basic} and $\name$. And for $\name$ we use $\hat{k} = k$. Different colors in each sub-figure mean different clusters.}
    \label{fig:cora_tsne}
\end{figure*}

In centralized settings, prior works~\cite{kumar2010clustering, awasthi2012improved} have achieved good performance both experimentally and theoretically.
\revision{They have proved that protocol~\ref{alg:centralized} can converge to the optimal clusters for the majority of nodes. Worth mentioning that these results pose no constraint on the data distribution.
}
By extending their framework to collaborative settings, we are able to provide a rigorous analysis of the performance of our method through Theorem~\ref{thm.1}.
Theorem~\ref{thm.1} illustrates that under certain conditions, Protocol~\ref{alg:collaborative_optimized} can accurately classify the majority of the nodes. 
Furthermore, we prove that our assumption serves as a compromise between the assumption in Definition~\ref{def.1} and that in~\cite{kumar2010clustering}.


Firstly, we introduce some notations (Table~\ref{tab:notation}) and definitions we use for our analysis. We use $\cluster$ as a cluster and $\mean(\cluster)$ as an operator to indicate the mean of the nodes assigned to $\cluster$. For consistency, we use $i, j$ as individual rows and columns of matrix $\X$ respectively, and use $l$ to indicate the index of $\Cli$, e.g., $\X_{i, j}$ is the $i^{th}$ row and $j^{th}$ column of $\X$ and $\X^l_i$ is the $i^{th}$ row of partitioned data $\X^l$. Lastly, $k$ denotes the number of clusters and $\lc$ denotes the number of local clusters.

Let $||\X||$ denote the spectral norm of a matrix $\X$, defined as $||\X||=max_{||u||=1}||\X u||_2$. 
Let $\mathcal{T}=\{T_1, T_2, \cdots, T_k\}$ be a set of fixed target clusters which is the optimal set of clusters one can achieve, we index clusters with $p, q$ and we use $\mean_q:=\mean(T_q)$ for simplicity.
We use $\X_q$ to denote the feature matrix of nodes indexed by $T_q$. Let $c(\X_i)$ denote the cluster index of $\X_i$. 
Let $C$ be a $n\times m$ matrix, and $C_i=\mean_{c(\X_i)}$ represent the centers of associated nodes. Note that we endow $\cluster$ and $T$ as different meanings, where $\cluster$ is the cluster that the algorithm outputs and $T$ is the optimal cluster.

Note that Protocol \ref{alg:collaborative_optimized} involves a local and a centralized $k$-means algorithm. To differentiate between these two algorithms, we propose the specific assumptions and conditions, which are outlined below.

For each participant $\Cli_l$ in the collaborative setting, Let $\mathcal{T}^l=\{T^l_1, T^l_2, \cdots, T^l_{\lc}\}$ be the local target clusters, we index local clusters with $r, s$. we assume that the feature matrix for $\Cli_l$ is denoted as $\X^l$. Similar to the prior definition, we use $\mean_r^l:=\mean(T^l_r)$ and let $\hat{C}^l$ be a $n \times m_l$ matrix of the local cluster means representing the local cluster centers of each node. 

For local cluster $u_r^l$ with $n_r = |T_r^l|$, define 
\begin{equation}\label{equation.delta}
    \Delta_r^l = \sqrt{\lc}\frac{||\X^l-\hat{C}^l||}{\sqrt{n_r}}.
\end{equation}
$Local\ center\ separation\ assumption$ states that for a large constant $c$, the means of any two local clusters $T^l_r$ and $T^l_s$ satisfy
\begin{equation}\label{eq.local_center_separation}
    ||\mean_r^l-\mean_s^l||\geq c(\Delta^l_r+\Delta^l_s).
\end{equation}

Intuitively, this assumption treats each participant as an independent and meaningful dataset for clustering. 
This is a reasonable expectation for more mutually beneficial cooperation in practice. 
For instance, TikTok and Twitter can build effective recommendation systems for their regular users independently. When aiming to benefit from each other's success, they may decide to cooperate in developing a more powerful recommendation system that integrates insights from both datasets.

\revision{Similarly, we define the $local\ proximity\ condition$ as follows, fix}
$i \in T_r^l$, if for any $s\neq r$, when projecting $\X^l_i$ onto the line connecting $\mean^l_r$ and $\mean^l_s$, the projection $\overline{\X}^l_i$ is closer to $\mean_r^l$ than to $\mean_s^l$ by at least $(\frac{1}{\sqrt{n_r}}+\frac{1}{\sqrt{n_s}})||\X^l-\hat{C}^l||$. The nodes satisfying $local\ proximity\ condition$ are called $local\ 1-good$, otherwise, we say the nodes are $local\ 1-bad$. 
\revision{To avoid confusion, the global versions of the proximity condition and the center separation assumption are formalized by treating local datasets as a unified whole and omitting the superscripts indicating party numbers.}

Now we define our proposed restricted proximity condition as follows.
\begin{definition}\label{def.2}
Let $\mathcal{T}=\{T_1, T_2, \cdots, T_k\}$ be the set of target clusters of the data $\X$, and $\hat{C}=[\hat{C}^1, \hat{C}^2, \cdots, \hat{C}^L]$. Fix $i\in T_q$  we say a node $\X_i$ satisfies the \textbf{restricted proximity condition} if for any $p\neq q$, 

(a)If the number of local 1-bad nodes is $\epsilon n$ for each participant, 
$\X_i$ is closer to $\mean(T_q)$ than to $\mean(T_p)$ by at least
\begin{equation}\label{eq.restricted_a}
    (\frac{1}{\sqrt{n_p}}+\frac{1}{\sqrt{n_q}})||\X-C||+\frac{2(1+O(\frac{1}{c})\sqrt{L}m_0)}{\sqrt{n_q}}||\X-\hat{C}||,    
\end{equation}
where $m_0=\sqrt{\frac{|T_q|}{\min_{l}|T^l_q|}}$. 

(b)If all nodes are local 1-good, 
$\X_i$ is closer to $\mean(T_q)$ than to $\mean(T_p)$ by at least 
\begin{equation}\label{eq.restricted_b}
    (\frac{1}{\sqrt{n_p}}+\frac{1}{\sqrt{n_q}})||\X-C||+\frac{2}{\sqrt{n_q}}||\X-\hat{C}||.
\end{equation}
\end{definition}

Intuitively, these notations make some assumptions about the data in collaborative settings. For (a) in our definition, it means that there can be at most $L\epsilon n$ $local\ 1-bad$ nodes in general. To capture the difference in $L$ participant, it adds $\frac{2(1+O(\frac{1}{c})\sqrt{L}m_0)}{\sqrt{n_q}}||X-\hat{C}||$ to the centralized condition $(\frac{1}{\sqrt{n_p}}+\frac{1}{\sqrt{n_q}})||\X-C||$, where $||\X-\hat{C}||$ captures the difficulty of clustering local data in each participant.

The proximity condition first proposed by \cite{kumar2010clustering} stated that the projection of $\X_i$ is closer to $\mean(U_q)$ than to $\mean(U_p)$ by at least $ck(\frac{1}{\sqrt{n_p}}+\frac{1}{\sqrt{n_q}})||\X-C||$. Later \cite{awasthi2012improved} improved this condition as Definition \ref{def.1} and still retrieve the target clusters.
\revision{When $\lc\geq k$, $||\X-\hat{C}|| \leq ||\X - C||$ since $\hat{C}$ is the local optimal centers.
}
For (a), $2(1+O(\frac{1}{c})\sqrt{L}m_0)<ck$ and for (b) $2<ck$ where $c>100$, 
the restricted proximity condition is stronger than Definition \ref{def.1}, but weaker than the proximity condition proposed by \cite{kumar2010clustering}.


\begin{theorem}\label{thm.1}
Let $\mathcal{T}=\{T_1, T_2, \cdots, T_k\}$ be the set of target clusters of the nodes $\X$. For participant $\Cli_l,1\leq l\leq L$, let $\mathcal{T}^l=\{T^l_1, T^l_2, \cdots, T^l_{\hat{k}}\}$ be the set of target clusters of local nodes $\X^l$.
Assume that each pair of clusters in $\mathcal{T}$ and $\mathcal{T}^l$ satisfy the (local) center separation assumption.
And all nodes $\X$ satisfy the restricted proximity condition, choosing the initial points which are better than a 10-approximation solution to the optimal clusters, if the number of local 1-bad nodes is $\epsilon n$ for each participant then
no more than $(L\epsilon+O(1)Lc^{-4})n$ nodes are misclassified. Moreover, if $\epsilon=0$, then all nodes are correctly assigned.
\end{theorem}
\revision{
\textit{Proof Sketch.} \textlabel{To}{R4:4:1} prove Theorem~\ref{thm.1}, we need to show that the proximity conditions in Definition~\ref{def.1} are satisfied for the nodes generated by the intersections of the local clusters in each participant. Then we can use Lemma~\ref{lemma.1} directly to achieve our results.
Note that we omitted SVD on these generated nodes, we only need to prove that the generated node is closer to $\mean_q$ than $\mean_p$ by at least 
$(\frac{1}{\sqrt{n_p}}+\frac{1}{\sqrt{n_q}})||\X-C||$. 
If the number of local 1-bad points in each participant is $\epsilon n$, according to Lemma~\ref{lemma.1} we have 
$||\mean(\hat{\cluster}^l_r)-\mean_r^l||\leq O(\frac{1}{c})\frac{1}{\sqrt{|T^l_r|}}||\X^l-\hat{C}^l||$. 
Note that the generated nodes are exactly the combinations of centers of local clusters and choose $i$ as the argmin of $||\X_i-\hat{\mean}_q$, following equation~\ref{equ.U-u} and equation~\ref{equ.indicator}, we have $||\X_i-\mean({\cluster}_q)\leq(\frac{1+O(\frac{1}{c})\sqrt{L}m_0}{\sqrt{n_q}})||X-\hat{C}||$.
Combining the triangle inequality and equation~\ref{eq.restricted_a}, each node satisfy the condition
$    ||{\mean(\hat{\cluster}_q)}-\mean(T_p)||-||{\mean(\hat{\cluster}_q)}-\mean(T_q)||
    \geq (\frac{1}{\sqrt{n_p}}+\frac{1}{\sqrt{n_q}})||\X-C||$. 
This concludes our proof.
}

\revision{The detailed proof is in Appendix~\ref{appdix.A}.
The proof indicates that if the initial conditions are reasonably favorable\footnote{\revision{The 10-approximation can be any t-approximation if c/t is large enough}}, our method can successfully classify the majority of nodes. This outcome is contingent upon the proportion of local nodes that meet the $local\ proximity\ condition$, which is $1-\epsilon$ in literature~\cite{awasthi2012improved, kumar2010clustering}, 
}

\revision{\textlabel{Figure}{tsne}
~\ref{fig:cora_tsne} shows the cluster results on ``Cora'' dataset using the t-SNE algorithm~\cite{t-SNE}. It experimentally shows that the cluster results of $\name$ are similar to the centralized method.
}

\revision{
\textlabel{The}{R1:3.2:1}
convergence speed of $\name$ depends on basic $\name$ and Protocol~\ref{alg:centralized} since it first runs Protocol~\ref{alg:centralized} locally then applies basic $\name$ on the centers of the intersections. Given that basic $\name$ is numerically identical to the centralized method, the convergence speed of $\name$ aligns with that of Protocol~\ref{alg:centralized}, which is the k-means method. This classic method has been proven to be effective and can converge within ten iterations experimentally~\cite{arthur2007k}.~\cite{har2005fast} has theoretically proven that k-means converges within a polynomial number of iterations ($\Omega(n)$).
In our experiments, we set the maximum number of iterations for $\name$ as 10. We find that it only needs about 4 iterations to converge for both $\name$ and the centralized protocol. This experimentally shows that $\name$'s convergence speed is comparable to the centralized method.
}

\begin{table}[ht]
\caption{Datasets Details}
\label{tab:datasets}
\centering
\begin{tabular}{|l|l|l|l|l|}
\hline
Dataset & Nodes & Edges & Features & Classes \\ \hline
Cora & 2708 & 5429 & 1433 & 7 \\
Citeseer & 3327 & 4732 & 3703 & 6 \\
Pubmed & 19717 & 44338 & 500 & 3\\
Wiki & 2405 & 17981 & 4973 & 17 \\ \hline
\end{tabular}
\end{table}

\begin{table*}[ht]
\caption{Summary of node classification accuracy (\%) results}
\label{tab:utility}
\centering
\begin{tabularx}{\textwidth}{XXXXXXXXX}
\hline
              & \multicolumn{4}{c}{\textbf{Cora}}                                         & \multicolumn{4}{c}{\textbf{Citeseer}}                                     \\ \cline{2-9} 
              & $L=2$            & $L=4$            & $L=8$            & $L=16$           & $L=2$            & $L=4$            & $L=8$            & $L=16$           \\ \hline
$\lc=k$       & 67.81$(\pm 1.78)$& 68.12$(\pm 0.33)$& 67.85$(\pm 1.01)$& 66.83$(\pm 1.71)$& 68.89$(\pm 0.12)$& 69.67$(\pm 0.02)$& 69.26$(\pm 0.01)$& 70.19$(\pm 0.55)$\\
$\lc=2k$      & 69.69$(\pm 0.95)$& $\mathbf{72.44}$$(\pm 0.51)$ & 71.28$(\pm 1.25)$& 71.97$(\pm 0.41)$& $\mathbf{69.74}(\pm 0.23)$ & 69.63$(\pm 0.23)$& 69.33$(\pm 0.06)$& $\mathbf{70.32}(\pm 0.28)$ \\
$\lc=4k$      & $\mathbf{72.12}(\pm 0.76)$ & 70.30$(\pm 1.60)$ & $\mathbf{71.89}(\pm 0.88)$& $\mathbf{72.78}(\pm 1.18)$ & 69.28$(\pm 0.14)$& $\mathbf{69.90}(\pm 0.18)$ & $\mathbf{69.96}(\pm 0.05)$ & 69.86$(\pm 0.16)$ \\
$\lc=8k$      & 70.49$(\pm 0.38)$ & 71.16$(\pm 2.06)$& 71.87$(\pm 1.69)$ & 70.96$(\pm 2.04)$& 69.60$(\pm 0.34)$& 69.71$(\pm 0.04)$& 69.83$(\pm 0.08)$& 69.84$(\pm 0.07)$ \\
b-$\name$ & 67.6$(\pm 0.1)$& 69.24$(\pm 1.14)$& 70.27$(\pm 1.04)$ & 69.63$(\pm 2.78)$& 68.29$(\pm 0.08)$& 68.66$(\pm 0.1)$    & 68.51$(\pm 0.08)$& 67.81$(\pm 0.48)$ \\
AGC           & \multicolumn{4}{c}{68.17 $(\pm 0.00)$}                                    & \multicolumn{4}{c}{68.40 $(\pm 0.01)$}                                    \\
GCC           & \multicolumn{4}{c}{74.29 $(\pm 0.00)$}                                    & \multicolumn{4}{c}{69.45 $(\pm 0.00)$}                                    \\ \hline
              & \multicolumn{4}{c}{\textbf{Pubmed}}                                       & \multicolumn{4}{c}{\textbf{Wiki}}                                         \\ \cline{2-9} 
              & $L=2$            & $L=4$            & $L=8$            & $L=16$           & $L=2$            & $L=4$            & $L=8$            & $L=16$           \\ \hline
$\lc=k$       & 69.67$(\pm 0.00)$ & 65.61$(\pm 5.43)$ & 69.67$(\pm 0.00)$ & 69.79$(\pm 0.00)$ & 45.72$(\pm 1.28)$ & 51.59$(\pm 1.93)$ & 52.78$(\pm 0.82)$   & 52.88$(\pm 0.14)$ \\
$\lc=2k$      & 69.71$(\pm 0.00)$& 66.14$(\pm 4.97)$& $\mathbf{69.85}(\pm 0.00)$ & $\mathbf{69.83}(\pm 0.01)$ & 49.63$(\pm 1.56)$& 51.16$(\pm 2.20)$& 53.46$(\pm 0.89)$ & $\mathbf{53.08}(\pm 0.65)$ \\
$\lc=4k$      & 69.71$(\pm 0.08)$& $\mathbf{69.85}(\pm 0.00)$ & 69.77$(\pm 0.00)$& 69.80$(\pm 0.02)$& $\mathbf{50.45}(\pm 0.27)$ & $\mathbf{53.57}(\pm 0.58)$ & 53.33$(\pm 1.28)$& 52.93$(\pm 0.78)$ \\
$\lc=8k$      & $\mathbf{69.72}(\pm 0.02)$ & 69.76$(\pm 0.07)$ & 69.67$(\pm 0.00)$& 69.81$(\pm 0.12)$& 50.35$(\pm 0.98)$  & 53.17$(\pm 0.73)$ & $\mathbf{53.54}(\pm 0.94)$ & 52.07$(\pm 0.85)$ \\
b-$\name$ & 69.69$(\pm 0)$& 69.77$(\pm 0)$& 69.74$(\pm 0)$ & 69.80$(\pm 0)$& 50.05$(\pm 1.84)$     & 50.92$(\pm 1.04)$    & 52.21$(\pm 0.54)$ & 52.02$(\pm 1.02)$  \\
AGC           & \multicolumn{4}{c}{69.78 $(\pm 0.00)$}                                    & \multicolumn{4}{c}{46.50 $(\pm 3.66)$}                                    \\
GCC           & \multicolumn{4}{c}{70.82 $(\pm 0.00)$}                                    & \multicolumn{4}{c}{54.56 $(\pm 0.00)$}                                    \\ \hline
\end{tabularx}
\end{table*}

\subsection{Security analysis}

\revision{
\textlabel{The}{R4:5:1}
privacy protection schemes for $\name$ are two-fold, incorporating both local data distortion methods and 
cryptographic algorithms to protect private data. Firstly, graph convolution is applied to the original local 
features, which will distort the original features. 
Subsequently, Singular Value Decomposition (SVD), which has been employed as an effective protection method in many works~\cite{polat2005svd, lakshmi2013svd, xu2006singular}, is applied to project the features to the top $\lc$ subspace. }

\revision{
\textlabel{For basic $\name$}{R1:3.3:1}, each participant calculates the distance between the cluster centers and nodes' features, updating the cluster centers locally, thereby mitigating any potential data leakage during these phases.
The potential information leakage primarily arises in two places,
\begin{itemize}[leftmargin=*]
    \item In line~14 and line~29, $\Cli_L$ tells other $\Cli_l$s the cluster assignment.
    \item In line~12 and line~27, all participants run secure aggregation to compute the distance for $Q$ iterations.
\end{itemize}
In the first place, the assignment will anyway be revealed at the end of a node classification task.
The second place will be protected by secure aggregation, ensuring the security of data during the communication phase. }

\revision{
\textlabel{It}{R4:5:2} has been proven by~\cite{bonawitz2017practical} that
although $\Cli_L$ can collude with $\tau-1$ passive participants, it still cannot learn anything beyond the sum of $L-\tau$ participants' inputs.  The theorems to prove this are provided in Appendix~\ref{appdix.B}. 
We then show that it is impossible for an adversary to deduce individual feature values from the aggregated sum. For a participant with \(n\) features, where \(n \geq 2\), this scenario can be reduced to solving \(n\) variables with 1 equation, which will result in an infinite number of solutions.
We also conducted experiments to demonstrate the extent of privacy leakage using real datasets. The outcomes of these experiments are presented in Appendix~\ref{appdix.B}.
}

\revision{For the security of $\name$}, we focus on analyzing Protocol~\ref{alg:collaborative_optimized} as 
the tree-based solution only changes the arrangement and will not reveal extra information.

{\bf $\Cli_l$'s security.} 
There are two places for potential information leakage:
\begin{itemize}[leftmargin=*]
\setlength{\itemsep}{0pt}
\setlength{\parsep}{0pt}
\setlength{\parskip}{0pt}
    \item In line 2 of Protocol~\ref{alg:collaborative_optimized}, $\Cli_l$ sends the nodes IDs in each cluster of $\{\hat{\cluster}_1^l, \cdots, \hat{\cluster}_{\hat{k}}^l\}$ to $\Cli_L$.
    \item In \revision{line~16, all $\Cli$s run Protocol~\ref{alg:collaborative_basic} on the virtual nodes $\mathbf{Y}$.}
\end{itemize}

The former only reveals the IDs in each of the $T$ local clusters.
However, the centers of the clusters are not revealed, hence the adversary learns little about the values.
Furthermore, the IDs in each cluster will be revealed at the end of $k$-means.
For the latter, \revision{\textlabel{compared}{R1:3.3:2} to the basic $\name$, $\name$ reveals only the sum of virtual nodes, which enhances the security in the aggregation part.}

{\bf $\Cli_L$'s security.}
There is one place for potential information leakage:
\begin{itemize}[leftmargin=*]
\setlength{\itemsep}{0pt}
\setlength{\parsep}{0pt}
\setlength{\parskip}{0pt}
    \item In line~10 of Protocol~\ref{alg:collaborative_optimized}, $\Cli_L$ sends the intersections to other $\Cli$s.
\end{itemize}
The intersections are calculated based on the node IDs received from other $\Cli$s. 
Therefore, the information leakage of $\Cli_L$ will clearly not be larger than $\Cli_i$ ($\Cli_i$'s security has been shown above).

\section{Implementation and Experiments}
\label{sec:experiments}

To systematically evaluate the utility and efficiency of our method, we pick four representative graph datasets: Cora, Citeseer, and Pubmed \cite{kipf2016variational} are citation networks where nodes are publications and are connected if one cites the other. Wiki \cite{yang2015network} is a webpage network where nodes represent webpages and are connected if one links to the other.

\subsection{Utility.} \label{subsec:utility}
\revisionminor{\textlabel{Since}{MinorR1:2:1} we are the first to explore this novel yet important vertical setting for unsupervised methods, there are no existing baselines. Therefore, to
prove the utility of our method, we selected basic $\name$ (b-$\name$) and the centralized methods $AGC$~\cite{zhang2019attributed} and GCC~\cite{fettal2022efficient} as baselines. Note that centralized methods do not face the data isolation problem, as the server has access to the whole dataset. To highlight the utility of our method in solving the data isolation problem, we aim to show that our method can achieve similar accuracy compared to the baselines. 
Specifically,} we compare the accuracy of $\name$ with AGC and \revision{basic~$\name$} with the same graph filter.
We set the order of graph filter $\psi$ as 9 for ``Cora'', 15 for ``Citeseer'', 60 for ``Pubmed'' and 2 for ``Wiki''. All experiments are repeated more than 5 times.
We set the number of local clusters ($\lc$) as a multiple of the number of final clusters ($k$). (cf. Table~\ref{tab:datasets})

Table~\ref{tab:utility} shows the accuracy achieved by our method with a different number of local clusters ($\lc$) and a different number of participants ($L$).
We highlighted the top results in our method for different numbers of participants in Table~\ref{tab:utility}

{\bf The results show that for the four datasets, our method can achieve the same level or even higher accuracy compared to the \revision{AGC and basic $\name$}.}
\revision{\textlabel{However}{R1:4:2}
, compared to GCC, our approach exhibits a lower performance on ``Cora'' and ``Wiki'' datasets, with a difference within $5\%$ in most cases, which underscores its competitive effectiveness despite the slight discrepancy.
}
Notice that for all the four datasets, the best results are achieved when the number of local clusters $\lc$ is set as $2k$ or $4k$ except for the ``Pubmed'' dataset with 2 participants $(\lc=8K)$ and ``Wiki'' dataset with 8 participants $(\lc=8k)$. Among different settings of $\lc$, $\lc=k$ tends to be the worst choice.
Intuitively, a larger number of local clusters $\lc$ means the size of the node space is larger, which will have more accurate results. 

\revisionminor{\textlabel{However}{MinorR2:1:1}, it may be counterintuitive that larger $\lc$ does not always guarantee better results. The reason lies in the theoretical analysis. When $\lc = k$, $||\X-\hat{C}|| \leq ||\X - C||$, and the \textbf{restricted proximity condition} is on the same scale of Definition~\ref{def.1}. Thus, the clustering results of $\name$ are close enough to the centralized method when $\lc\geq k$. }

\revisionminor{
Nevertheless, both the $\textbf{restricted proximity condition}$ and $Local\ center\ separation\ assumption$ are non-monotonic with respect to $\lc$.
As $\lc$ increases, the size of local clusters becomes smaller. Consequently, $m_0^2$ in~\ref{eq.restricted_a}, which is inversely proportional to the minimal size of local clusters, will become larger. Additionally, $\Delta$, as defined in equation~\ref{equation.delta}, is positively correlated with $k/n_q$, where $n_q$ is the size of the local cluster indexed by $q$. These two factors limit the applicability of  $\textbf{restricted proximity condition}$ and \textbf{proximity condition}, making them more difficult to satisfy. 
Therefore, $\lc=2k$ or $4k$ usually yields more accurate results compared to $\lc=8k$.
}

On the other hand, larger $\lc$ will pose more pressure on the communication budget because we need to communicate the intersections.
Recall that our intuition is to use local information to reduce the sample space such that we can avoid a lot of expensive cryptographic operations. When $L=2$, we need $\lc\leq k$ for ``Cora'' and ``Citeseer'', and $\lc\leq4k$ for ``Pubmed'' to ensure that the number of intersections for communication is at most one-tenth the number of nodes $n$. For ``Wiki'', when $\lc\leq k$, there will be $0.12n$ intersections for communication.
However, when we choose $\lc=k$, $\name$ can still achieve comparable accuracy compared to the baselines.
From this perspective, our method is more suitable for large-scale datasets like ``Pubmed'', whose $n$ can be tens of thousands or even hundreds of thousands. 

Moreover, we find that different numbers of participants do not significantly affect the results. For the ``Wiki'' dataset, an increase in participants results in decreased accuracy. In contrast, for the ``Citeseer'' dataset, accuracy improves with a greater number of participants. Nonetheless, our methods can achieve similar accuracy for all datasets compared to the baselines when setting an appropriate $\lc$.

These results empirically show that in our restricted proximity condition ((~\ref{eq.restricted_a}) and((~\ref{eq.restricted_b})), the factor $O(\frac{1}{c})$ can make the second term small enough such that only the first term plays a vital role in experiments, which is similar to the centralized setting.

\begin{table}[ht]
\centering
\caption{Results of kCAGC (\(\hat{k}=k\)) compared to GraphSAGE trained on isolated data and mixed data on the Cora and Pubmed datasets.}
\label{tab:kCAGC_vs_GraphSage}
\begin{tabular}{lcccc}
\hline
 & \multicolumn{2}{c}{Cora} & \multicolumn{2}{c}{Pubmed} \\
\cmidrule(r){2-3}\cmidrule(r){4-5}
 & Acc & F1 & Acc & F1 \\
\hline
$\name$ & \textbf{67.81} & \textbf{61.83} & \textbf{69.67} & \textbf{68.51} \\
 & ($\pm 0.17$) & ($\pm 0.09$) & ($\pm 0.00$) & ($\pm 0.00$) \\
GraphSAGE$_A$ & 55.00 & 43.80 & 50.28 & 35.40 \\
 & ($\pm 0.00$) & ($\pm 0.00$) & ($\pm 0.00$) & ($\pm 0.00$) \\
GraphSAGE$_B$ & 55.87 & 46.94 & 59.03 & 48.13 \\
 & ($\pm 0.00$) & ($\pm 0.00$) & ($\pm 0.00$) & ($\pm 0.00$) \\
GraphSAGE$_{A+B}$ & 71.80 & 68.08 & 66.11 & 62.50 \\
 & ($\pm 0.00$) & ($\pm 0.00$) & ($\pm 0.00$) & ($\pm 0.00$) \\
\hline
\end{tabular}
\end{table}

\revision{
\textlabel{To}{sage} 
demonstrate that $\name$ effectively addresses the data isolation problem, we compare $\name$ with the semi-supervised version of GraphSAGE models\footnote{We use the pytorch implementation from \url{https://github.com/zhao-tong/GNNs-easy-to-use}. The GraphSAGE embeddings are unsupervised, but the classifier is trained supervised, therefore it is semi-supervised.} trained on isolated data. We partition the dataset into two non-overlapping subsets, $A$ and $B$, and train the models on these separate subsets while maintaining the same graph structure. The comparative results are presented in Table~\ref{tab:kCAGC_vs_GraphSage}. We highlight the highest accuracy and F1-scores for both $\name$ and the models trained on the isolated data in $\textbf{bold}$.
Our findings consistently show that $\name$ surpasses the performance of models trained on isolated data, even with a small $\lc$. This empirically validates that $\name$ is capable of effectively mitigating the challenges posed by data isolation. Furthermore, we notice that even for a semi-supervised model trained on the whole data, our methods can achieve comparable cluster results.
}

\revision{
We also conducted experiments to evaluate our methods on a different graph filter in Appendix~\ref{appdix.add_exp}.
}
\begin{table*}[ht]
\caption{Efficiency of k-CAGC in LAN.}
\label{tab:all_efficient}
\centering
\begin{tabular}{lcccccccccccccccc}
\hline
 & \multicolumn{16}{c}{Training Time (s)} \\
\cmidrule(lr){2-17}
 & \multicolumn{4}{c}{Cora} & \multicolumn{4}{c}{Citeseer} & \multicolumn{4}{c}{Pubmed} & \multicolumn{4}{c}{Wiki} \\ 
\cmidrule(lr){2-5} \cmidrule(lr){6-9} \cmidrule(lr){10-13} \cmidrule(lr){14-17}
$L$ & 2 & 4 & 8 & 16 & 2 & 4 & 8 & 16 & 2 & 4 & 8 & 16 & 2 & 4 & 8 & 16 \\ 
\hline
$\lc=k$ & 1.16 & 3.43 & 7.86 & 16.67 & 1.15 & 3.31 & 7.69 & 16.39 & 1.17 & 3.23 & 7.26 & 15.57 & 2.23 & 5.09 & 10.56 & 19.92 \\
$\lc=2k$ & 1.47 & 4.28 & 9.40 & 19.10 & 1.50 & 3.97 & 9.04 & 18.49 & 1.82 & 3.81 & 7.93 & 16.70 & 3.81 & 7.40 & 14.01 & 26.91 \\
$\lc=4k$ & 2.49 & 5.80 & 11.52 & 22.56 & 2.28 & 5.42 & 10.94 & 20.84 & 3.03 & 5.45 & 10.43 & 19.93 & 9.78 & 21.70 & 43.04 & 92.25 \\
$\lc=8k$ & 5.18 & 12.62 & 26.25 & 53.27 & 4.23 & 9.69 & 19.89 & 40.38 & 5.88 & 9.18 & 14.00 & 24.17 & 58.52 & 153.40 & 326.42 & 753.64 \\
\hline
\end{tabular}
\end{table*}

\begin{table}[ht]
\centering
\caption{Time efficiency for centralized AGC on various datasets.}
\label{tab:time_efficiency_datasets}
\begin{tabular}{|l|l|l|l|l|}
\hline
 & Cora & Citeseer & Pubmed & Wiki \\
\hline
Time (s) & 0.13 $\pm$ 0.02 & 0.13 $\pm$ 0.02 & 0.14 $\pm$ 0.02 & 0.16 $\pm$ 0.02 \\
\hline
\end{tabular}
\end{table}

\subsection{Efficiency.}

We fully implement our method in C++ using GMP\footnote{\url{https://gmplib.org/}} for cryptographic operations.
We deploy our implementation on a machine that contains 48 2.50GHz CPUs, 128 GB memory; we spawn up to 16 processes, and each process (including the baseline) runs as a single participant. We conduct our method on two network settings, LAN of 1000mbps bandwidth with no decay and WAN of 400mbps bandwidth with 100ms decay. The results for WAN settings are in Appendix~\ref{appendix.E3} Unfortunately, we find that current collaborative methods pay little attention to the efficiency in collaborative settings. We show the time consumption for $\name$ as a baseline in collaborative settings.

Table~\ref{tab:all_efficient} and Table~\ref{tab:all_efficient_2} show the training time of $\name$ with different $\lc$ and different $L$.
Table~\ref{tab:all_efficient} shows the training time of $k-CAGC$ for the four datasets in LAN. With the same $\lc$ and $L$, the training time for ``Cora'', ``Citeseer'' and ``Pubmed''  is similar. 
The training time for these three datasets is within $1$ minute for all settings of $L$ and $\lc$.
When $L=2,4$ and $\lc = k,2k$, the training time is within $10$ seconds.
For the ``Wiki'' dataset, when $L=16$ and $\lc=8k$, the training time is about $13$ minutes.


The results indicate that the training time for our method is not affected by the number of nodes in the graph, but rather depends on the number of participants and the number of local clusters. 
For the "Pubmed" dataset, which has nearly ten times as many nodes as the other three datasets, the training time is the shortest. For the "Wiki" dataset with 17 classes, training times are within 30 seconds for $\lc=k,2k$, and $L\leq16$, but it increases to almost 13 minutes when $\lc=8k$ and $L=16$. This is because increasing the number of local clusters or participants raises the time consumed by the intersection process and hence, results in longer training times. As discussed in Section~\ref{subsec.optimized_method}, our method requires $O((Q+1)L\cdot\lc^3)$ secure aggregations, so smaller $\lc$ values should be selected for better efficiency. Nonetheless, our method performs well for all $\lc$ values, and we can set $\lc$ as $\lc=k,2k$ for all experiments to achieve optimal efficiency.

Table~\ref{tab:all_efficient_2} shows the training time in WAN. The results are similar to LAN. Given 100ms decay in WAN, the minimum training time for the four datasets is about $10$ seconds. And for ``Wiki'' with $\lc=8K, L=16$, the training time is about $20$ minutes. However, when $\lc = k,2k$, the training time for all experiments will be within $1.5$ minutes.

\revision{\textlabel{Table}{R4:6:1}
~\ref{tab:time_efficiency_datasets} shows the training time of centralized AGC models which are usually within 1 second. It shows that collaborative learning can substantially affect time efficiency. Therefore, the efficiency of collaborative learning is crucial. 
}

\section{Conclusion}


In this paper, we proposed a novel approach for collaboratively attributed graph clustering based on the $k$-means algorithm. Our approach can effectively handle vertically partitioned data by utilizing graph filters to obtain filtered feature sets from each participant's local data. We provided a theoretical analysis to demonstrate the correctness of our method and showed that the success of local clustering on each participant's data contributes to the overall collaboration.

\revision{
Our method is as efficient as centralized approaches, completing clustering in minutes. However, like $k$-means, it depends on initial points, which can lead to inaccuracies. To mitigate potential societal impacts, we plan to enhance robustness by optimizing initialize strategies in future work.
}

\bibliographystyle{IEEEtran}
\bibliography{IEEEabrv,references}
\vspace{-10 mm}
\begin{IEEEbiography}[{\includegraphics[width=1in,height=1.25in,clip,keepaspectratio]{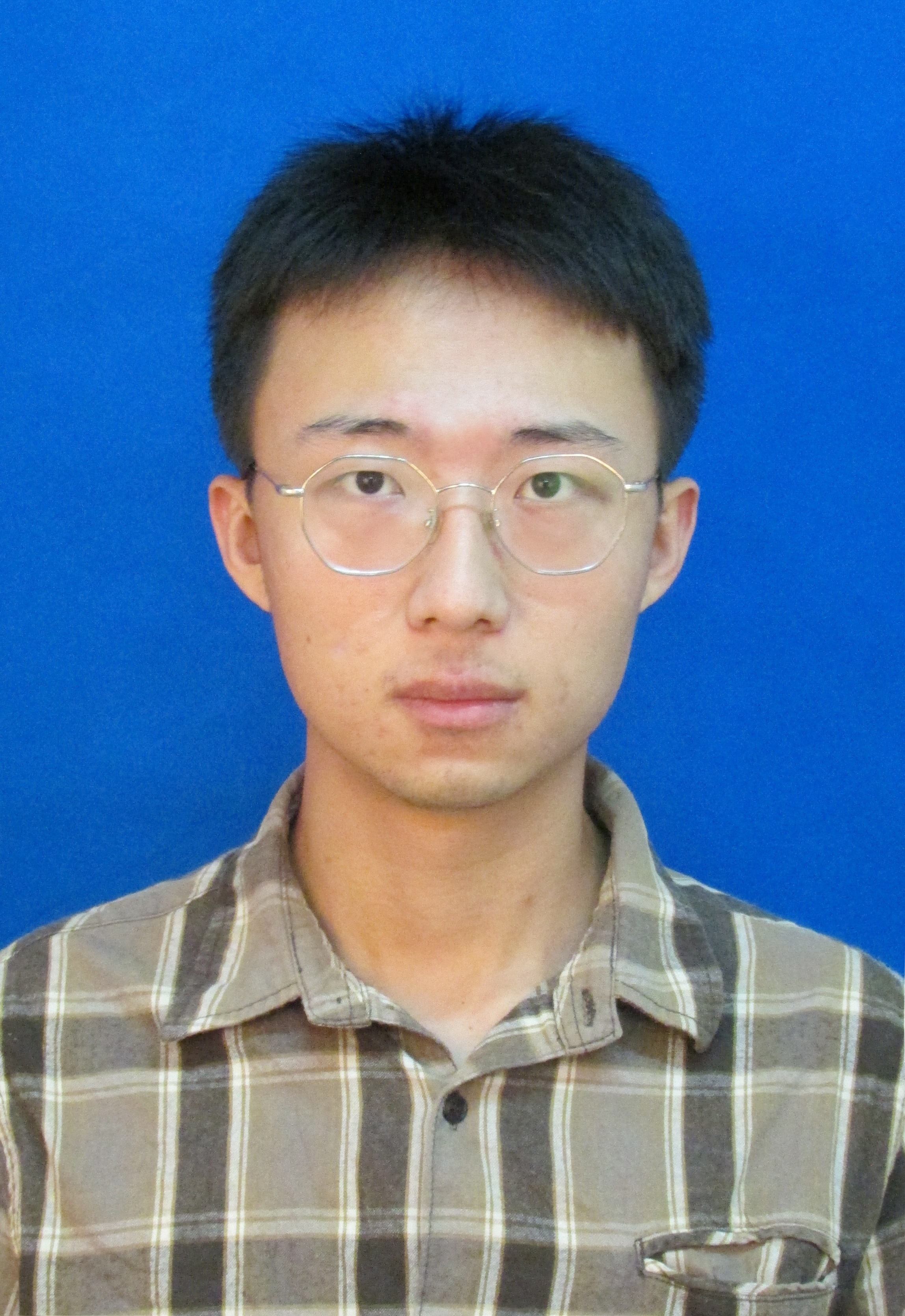}}]{Zhang Rui} graduated from the School of Mathematics in Zhejiang University, Hangzhou China, in 2020. Currently, he is working on his PhD in the School of Mathematics of Zhejiang University, Hangzhou China. His research interests include machine learning, federated learning, and adversarial training algorithms.
\end{IEEEbiography}
\vspace{-10 mm}
\begin{IEEEbiography}[{\includegraphics[width=1in,height=1.25in,clip,keepaspectratio]{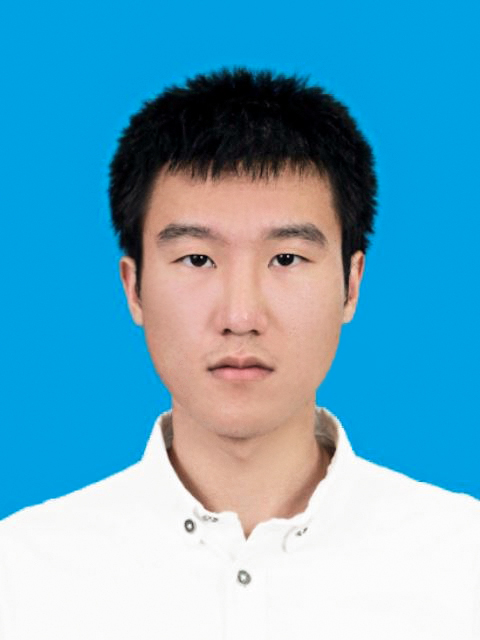}}]{Xiaoyang Hou} is a Ph.D. student at Zhejiang University. Before that, he received his bachelor's degree in 2021 from Zhejiang University of Technology. His research is on federated learning and applied cryptography.
\end{IEEEbiography}
\vspace{-10 mm}
\begin{IEEEbiography}[{\includegraphics[width=1in,height=1.25in,clip,keepaspectratio]{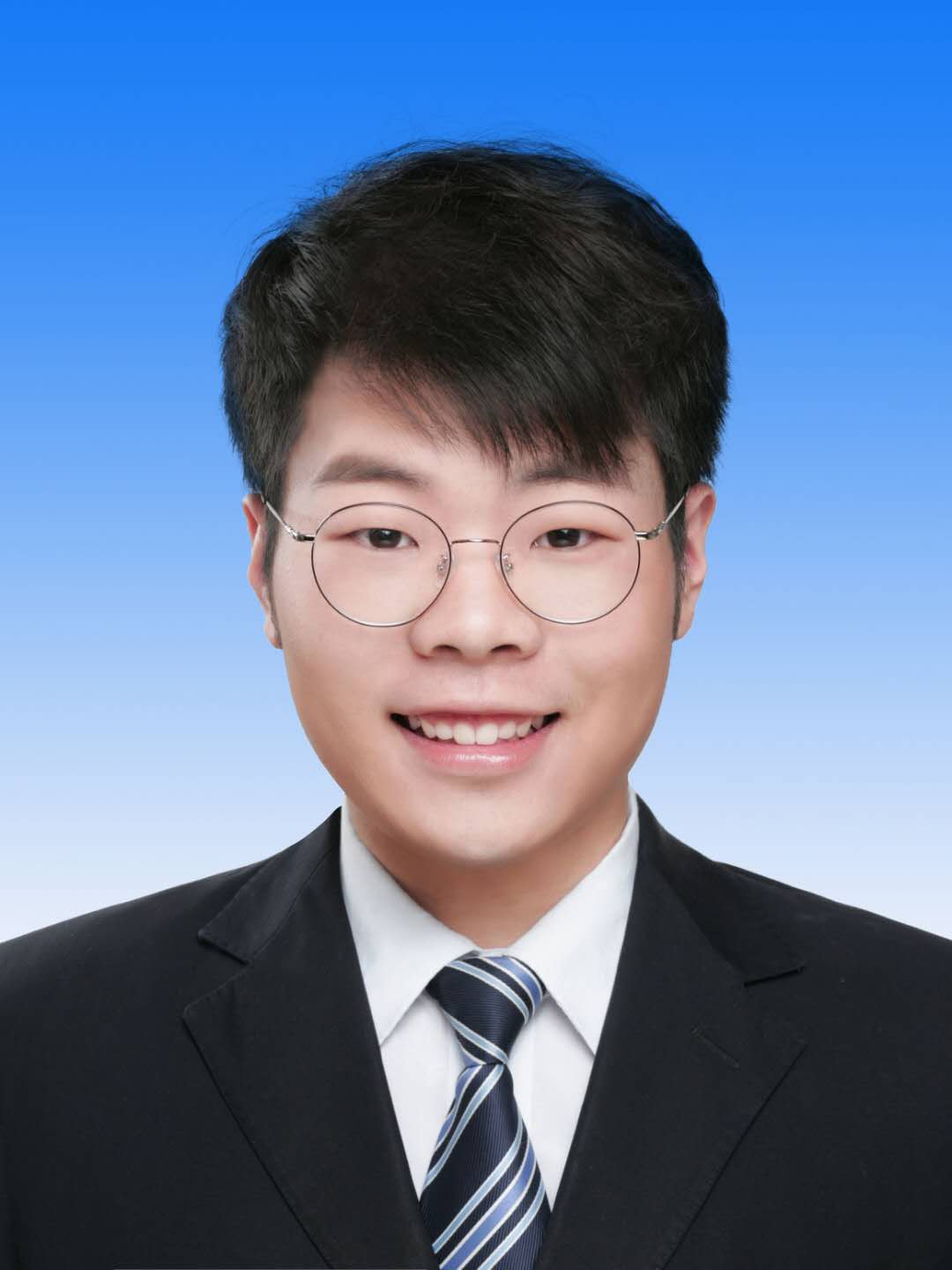}}]{Zhihua Tian} received the B.Sc. degree in stastics from Shandong University, Ji’nan, China, in 2020. Currently, he is pursuing a Ph.D. degree in the School of Cyber Science and Technology of Zhejiang University, Hangzhou, China. His
research interests include machine learning, federated learning, and adversarial training algorithms. 
\end{IEEEbiography}
\vspace{-10 mm}
\begin{IEEEbiography}[{\includegraphics[width=1in,height=1.25in,clip,keepaspectratio]{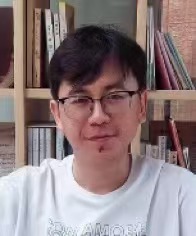}}]{Enchao Gong} Graduated with a master’s degree from Shanghai Normal University in 2012 and is currently engaged in anti-money laundering work at Ant Group, with some research experience in graph computing and algorithms.
\end{IEEEbiography}
\vspace{-10 mm}
\begin{IEEEbiography}[{\includegraphics[width=1in,height=1.25in,clip,keepaspectratio]{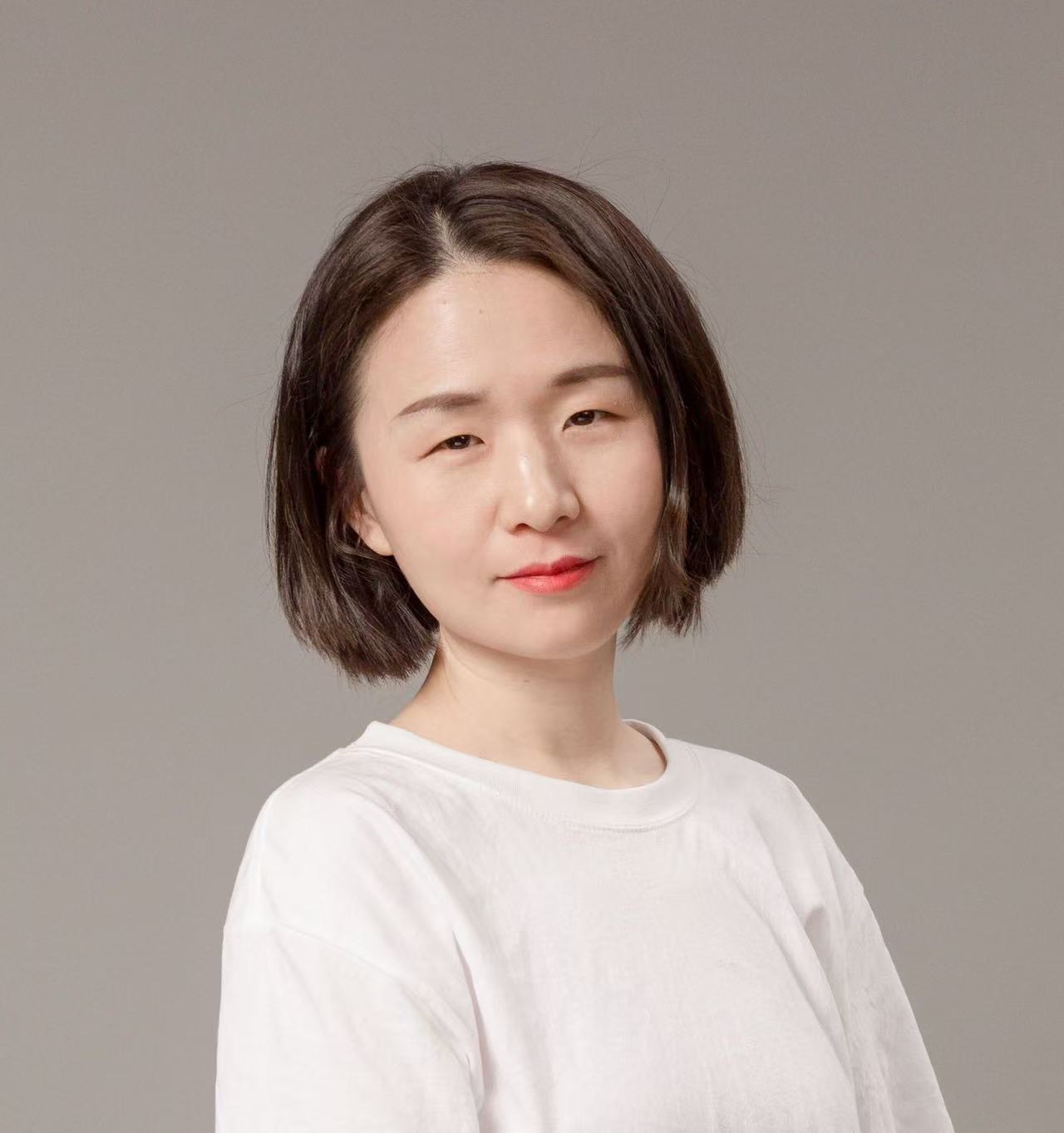}}]{He Yan} graduated from School of  Computer Science and Technology in Huazhong University of Science and Technology, WuHan China, in 2012. Currently, she is working at AntGroup now. She is engaged in the technologies related to anti-money laundering.
\end{IEEEbiography}
\vspace{-10 mm}
\begin{IEEEbiography}[{\includegraphics[width=1in,height=1.25in,clip,keepaspectratio]{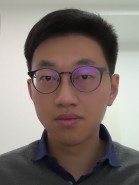}}]{Jian Liu}
is a ZJU100 Young Professor at Zhejiang University. Before that, he was a postdoctoral researcher at UC Berkeley. He got his PhD in July 2018 from Aalto University. His research is on Applied Cryptography, Distributed Systems, Blockchains, and Machine Learning. He is interested in building real-world systems that are provably secure, easy to use, and inexpensive to deploy.
\end{IEEEbiography}
\vspace{-10 mm}
\begin{IEEEbiography}[{\includegraphics[width=1in,height=1.25in,clip,keepaspectratio]{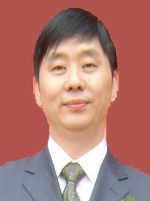}}]{Qingbiao Wu}
is a doctoral supervisor at Zhejiang University, currently serving as the director of the Institute of Science and Engineering Computing and deputy director of the North Star Aerospace Innovation Design Center. He also serves as the chairman of the Zhejiang Provincial Applied Mathematics Society and director of various academic associations in China. His research includes big data analysis, machine learning and numerical methods. 
\end{IEEEbiography}
\vspace{-10 mm}
\begin{IEEEbiography}[{\includegraphics[width=1in,height=1.25in,clip,keepaspectratio]{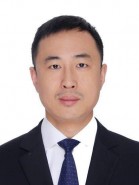}}]{Kui Ren}
is a Professor and Associate Dean of the College of Computer Science and Technology at Zhejiang University, where he also directs the Institute of Cyber Science and Technology. Before that, he was SUNY Empire Innovation Professor at the State University of New York at Buffalo, USA. He received his PhD degree in Electrical and Computer Engineering from Worcester technic Institute. Kui’s current research interests include Data Security, IoT Security, AI Security, and Privacy. Kui has published extensively in peer-reviewed journals and conferences and received the Test-of-time Paper Award from IEEE INFOCOM and many Best Paper Awards, including ACM MobiSys, IEEE ICDCS, IEEE ICNP, IEEE Globecom, ACM/IEEE IWQoS, etc. Kui is a Fellow of ACM and IEEE. He is a frequent reviewer for funding agencies internationally and serves on the editorial boards of many IEEE and ACM journals. Among others, he currently serves as Chair of SIGSAC ofACM China Council, a member of ACM ASIACCS steering committee, and a member of S\&T Committee of Ministry of Education of China.
\end{IEEEbiography}

\clearpage
\appendices
\section{List of Notations}

\begin{table}[ht]
\caption{List of important notations}
\label{tab:notation}
\centering
\begin{tabular}{l|l}
\hline
$\textbf{Symbols}$  & \textbf{Description}\\\hline
$\Cli^l$         & the $l^{th}$ participant                         \\\hline
$\X$             & feature matrix                                   \\\hline
$\hat{\X}$       & the projection of $\X$ on the top $k$ singular vectors.                                                          \\\hline
$\X_i$           & the $i^{th}$ node                                  \\\hline
$\X^l$          & the local data of $\Cli^l$                        \\\hline
$\X_i^j$         & the $i^{th}$ row, $j^{th}$ column of $\X$        \\\hline
$\overline{\X_i}$ & the projection of the $i^{th}$ node onto the line.\\\hline
$\mathcal{G}$    & an attributed non-direct graph                   \\\hline

$G$              & the graph filter to get the node embeddings      \\\hline
$\mathcal{U}$    & the set of clusters output by our methods        \\\hline
$\cluster$       & a cluster                                        \\\hline
$\hat{\cluster}$ & a local cluster                                  \\\hline
$S_r/S_q$        & the set of intermediate clusters during training \\\hline
$\mathcal{T}$    & the set of target clusters                       \\\hline
$T$              & a target cluster                                 \\\hline
$T^l$            & a target cluster of $\Cli^l$                     \\\hline
$\mean$          & a center(mean of corresponding cluster)          \\\hline
$\mean^l$        & a center of the local clusters of $\Cli^l$       \\\hline
$C$              & a matrix composed of the centers of each node    \\\hline
$\hat{C}^l$      & a matrix composed of the local centers of $\Cli^l$    \\\hline
$\hat{C}$        & a matrix composed of  $\hat{C}^l$ of each $\Cli^l$     \\\hline
$A$              & the adjacency matrix of the graph                \\\hline
$L_s$            & the symmetric normalized Laplacian               \\\hline
$L$              & the number of participants                       \\\hline
$k$              & the number of clusters                           \\\hline
$\lc$            & the number of local clusters                     \\\hline
$\psi$           & the order of the filter                          \\\hline
$n$              & the number of nodes                              \\ \hline
\end{tabular}
\end{table}

\section{Proof for Theorem~\ref{thm.1}}
\label{appdix.A}
We first introduce some conditions we will use to prove our theorem.
For cluster $T_q$ with $n_q = |T_q|$, we define
\begin{equation}
    \Delta_q = \frac{1}{\sqrt{n_q}}\sqrt{k}||\X-C||.
\end{equation}
$Center\ separation$ assumes that for a large constant $c$\footnote{$c>100$ will be sufficient}, the means of any two clusters $T_p$ and $T_q$ satisfy
\begin{equation}\label{eq.center_separation}
    ||\mean_p-\mean_q||\geq c(\Delta_p+\Delta_q).    
\end{equation}

With $proximity\  condition$ defined in Definition \ref{def.1} \cite{awasthi2012improved} and the above assumption, Lemma~\ref{lemma.1} \cite{awasthi2012improved} shows that for a set of fixed target clusters, Protocol \ref{alg:centralized} correctly clusters all but a small fraction of the nodes.
\begin{definition}\label{def.1}
Fix $i \in T_q$, we say a node $\X_i$ satisfies the \textbf{proximity condition} if for any $p\neq q$, when projecting $\X_i$ onto the line connecting $\mean_p$ and $\mean_q$, the projection of $\X_i$ is closer to $\mean_q$ than to $\mean_p$ by at least 
\begin{equation} \label{eq.pro_cond}
    (\frac{1}{\sqrt{n_p}}+\frac{1}{\sqrt{n_q}})||\X-C||.
\end{equation}
The nodes satisfying the proximity condition are called $1-good$, otherwise we say the nodes are $1-bad$.
\end{definition}
Here we introduce Lemma~\ref{lemma.1} from \cite{awasthi2012improved} similar to the lemma introduced by  \cite{dennis2021heterogeneity}.
\begin{lemma}\label{lemma.1}(Awasthi-Sheffet,2011)\cite{awasthi2012improved}. Let $\mathcal{T}=\{T_1, \cdots, T_k\}$ be the target clusters. Assume that each pair of clusters $T_r$ and $T_s$ satisfy the center separation assumption. Then after constructing clusters $S_r$ in Protocol \ref{alg:centralized}, for every r, it holds that
\begin{equation}
    ||\mean(S_r)-\mean_r||\leq O(\frac{1}{c})\frac{1}{\sqrt{n_r}}||\X-C||.
\end{equation}
 If the number of $1-bad$ nodes is $\epsilon n$, then (a) the clusters $\{\cluster_1, \cluster_2, \cdots, \cluster_k \}$ misclassify no more than $(\epsilon + O(1)c^{-4})n$ nodes and (b) $\epsilon<O((c-\frac{1}{\sqrt{k}})^{-2})$. Finally, if $\epsilon=0$ then all nodes are correctly assigned, and Protocol \ref{alg:centralized} converges to the true centers.
\end{lemma}
Prior works show that Protocol~\ref{alg:centralized} can converge to the optimal clusters for most nodes. Worth mentioning that distance the proximity condition is scaled up by $k$ factor in Kumar and Kannan~\cite{kumar2010clustering}. 
The results pose no constraint on the distribution of data.

Now, we prove Theorem~1.
\begin{proof}
For each participant $\Cli_l$, if the number of local 1-bad points is $\epsilon n$, Lemma 1 holds and we get that no more than $(\epsilon+O(1)c^{-4})n$ points are misclassified.
and $||\mean(\hat{\cluster}^l_r)-\mean^l_r||\leq O(\frac{1}{c})\frac{1}{\sqrt{|T^l_r|}}||\X^l-\hat{C}^l||$.

Let $\hat{\mean}_q$ denote the center of new point assigned to $T_q$ that generated by the intersections of $T^l, l=1,2, \cdots, L,$ and let $\hat{\cluster}_q$ denote the intersections of $\hat{\cluster}^l$ assigned to $T_q$. To be specific, for $\Cli_l$, if $\hat{\mean}_q=[\dots, \mean_r^l, \dots]$, we define $\mean^l_q:=\mean_r^l$. And $\mean(\hat{\cluster}_q^l)$ is similarly defined. Since 
\begin{equation}\label{equ.U-u}
    \begin{aligned}
    &||\mean(\hat{\cluster}_q)-\hat{\mean}_q||\\
    =&||[\mean(\hat{\cluster}^1_q)-\mean^1_q, \cdots, \mean(\hat{\cluster}^L_q)-\mean^L_q]||\\
    =&(||\mean(\hat{\cluster}^1_q)-\mean^l_q||^2+\cdots+||\mean(\hat{\cluster}^L_q)-\mean^L_q||^2)^{\frac{1}{2}}\\
    \leq&O(\frac{1}{c})\frac{1}{\min_{l}\sqrt{|T^l_q|}}(||\X^1-\hat{C}^1||^2+\cdots\\&+||\X^L-\hat{C}^L||^2)^{\frac{1}{2}}\\
    \leq&O(\frac{1}{c})\frac{\sqrt{L}m_0}{\sqrt{n_q}}||\X-\hat{C}||.
    \end{aligned}
\end{equation}

Since we have used SVD on the local data to get the projections, $\X = \overline{\X}$.
According to the restricted proximity assumption that for $\forall i \in T_q$, for any $p\neq q$,
\begin{equation}
    \begin{aligned}
    &||{\X_i}-\mean(T_p)||-||{\X_i}-\mean(T_q)||\geq \\
    &(\frac{1}{\sqrt{n_p}}+\frac{1}{\sqrt{n_q}})||\X-C||+ \frac{2(1+O(\frac{1}{c})\sqrt{L}m_0)}{\sqrt{n_q}}||X-\hat{C}||.
    \end{aligned}
\end{equation}
For $i = argmin _i ||\X_i-\hat{\mean}_q||$, let $u$ be an indicator vector for points in $T_q$,

\begin{equation}\label{equ.indicator}
    \begin{aligned}
|||T_q|(\X_i-\hat{\mean}_q)||&\leq||(\X-\hat{C})\cdot u||\\
&\leq||\X-\hat{C}||||u||,\\
||\X_i-\hat{\mean}_q||
&\leq \frac{1}{\sqrt{n_q}}||\X-\hat{C}||.
    \end{aligned}
\end{equation}

\begin{equation}
    \begin{aligned}
    ||\X_i-\mean(\hat{\cluster}_q)||\leq&||X_i-\hat{\mean}_q||+||\mean(\hat{\cluster}_q)-\hat{\mean}_q||\\
    \leq&(\frac{1+O(\frac{1}{c})\sqrt{L}m_0}{\sqrt{n_q}})||X-\hat{C}||.
    \end{aligned}
\end{equation}

\begin{equation}
    \begin{aligned}
    &\quad\  ||{\mean(\hat{\cluster}_q)}-\mean(T_p)||-||{\mean(\hat{\cluster}_q)}-\mean(T_q)||\\
    &\geq ||{\X_i}-\mean(T_p)||-||{\X_i}-\mean(T_q)||- 2||\mean(\hat{\cluster}_q)-\X_i||\\
    &\geq (\frac{1}{\sqrt{n_p}}+\frac{1}{\sqrt{n_q}})||\X-C||
    .
    \end{aligned}
\end{equation}
If all local data points are local 1-good, then $\mean(\hat{\cluster}_q)=\mean_q$. Inequality (22) can be similarly achieved.
Then all the new points generated by the intersections satisfying the proximity condition, Lemma \ref{lemma.1} can be used once the initial points in Protocol \ref{alg:collaborative_optimized} are close to the true clusters (any 10-approximation solution will do \cite{kumar2010clustering}).
This concludes our proof.
\end{proof}



\section{Secure Aggregation}\label{appdix.B}
\revision{
The secure aggregation is proven to be secure under our honest but curious setting, regardless of how and when parties abort. Assuming that a server S interacts with a set $\mathcal{U}$ of $n$
users (denoted with logical identities $1, \cdots, n$)
and the threshold is set to $t$. we denote with $\mathcal{U}_i$ 
the subset of the users that correctly sent their message to the 
server at round $i-1$, such that
$U \supseteq \mathcal{U}_1 \supseteq \mathcal{U}_2 \supseteq \mathcal{U}_3 \supseteq \mathcal{U}_4 \supseteq \mathcal{U}_5$. }

\revision{
Given any subset $\mathcal{C} \subseteq \mathcal{U} \cup {S}$ of the parties, let
$REAL^{\mathcal{U}, t, k}_\mathcal{C}
(x_\mathcal{U}, \mathcal{U}_1, \mathcal{U}_2, \mathcal{U}_3, \mathcal{U}_4, \mathcal{U}_5)$ be a random variable representing the combined views of all parties in $\mathcal{C}$, where $k$ is a security parameter. The following theorems shows that the joint view of any subset of honest users (excluding the server) can be simulated given only the knowledge of the inputs of those users. Intuitively, this means that those users learn “nothing more” than their own inputs.
The first theorem provides the guarantee where the server is semi-honest, and the second theorem proves the security where the semi-honest server can collude with some honest but curious client.}

\begin{theorem}[Theorem 6.2 in~\cite{bonawitz2017practical}]\label{thm.security1}
\revision{
(Honest But Curious Security, against clients
only). 
There exists a PPT simulator SIM such that for all k, t,$\mathcal{U}$
with 
$t \leq |\mathcal{U}|, x_\mathcal{U}, \mathcal{U}_1, \mathcal{U}_2, \mathcal{U}_3, \mathcal{U}_4, \mathcal{U}_5$ 
and $\mathcal{C}$ such that $\mathcal{C} \subseteq \mathcal{U}$,
$\mathcal{U} \supseteq \mathcal{U}_1 \supseteq \mathcal{U}_2 \supseteq \mathcal{U}_3 \supseteq \mathcal{U}_4 \supseteq \mathcal{U}_5$, the output of SIM is perfectly
indistinguishable from the output of 
$REAL^{\mathcal{U}, t, k}_\mathcal{C}
:$
\begin{equation*}
    \begin{aligned}
    REAL^{\mathcal{U}, t, k}_\mathcal{C}
    (x_\mathcal{U}, \mathcal{U}_1, \mathcal{U}_2, \mathcal{U}_3, \mathcal{U}_4, \mathcal{U}_5)\\
    \equiv
    SIM^{\mathcal{U}, t, k}_\mathcal{C}
    (x_\mathcal{C}, \mathcal{U}_1, \mathcal{U}_2, \mathcal{U}_3, \mathcal{U}_4, \mathcal{U}_5).  
    \end{aligned}
\end{equation*}
}
\end{theorem}

\begin{theorem}[Theorem 6.3 in~\cite{bonawitz2017practical}]\label{thm.security2}
\revision{(Honest But Curious Security, against clients
only). 
There exists a PPT simulator SIM such that for all t,$\mathcal{U}$
$, x_\mathcal{U}, \mathcal{U}_1, \mathcal{U}_2, \mathcal{U}_3, \mathcal{U}_4$ 
and $\mathcal{C}$ such that $\mathcal{C} \subseteq \mathcal{U}\cup\{S\}$, $|\mathcal{C}\backslash \{S\}|<t$,
$\mathcal{U} \supseteq \mathcal{U}_1 \supseteq \mathcal{U}_2 \supseteq \mathcal{U}_3 \supseteq \mathcal{U}_4 \supseteq \mathcal{U}_5$, the output of SIM is computationally
indistinguishable from the output of 
$REAL^{\mathcal{U}, t, k}_\mathcal{C}
:$
\begin{equation*}
    \begin{aligned}
    REAL^{\mathcal{U}, t, k}_\mathcal{C}
    (x_\mathcal{U}, \mathcal{U}_1, \mathcal{U}_2, \mathcal{U}_3, \mathcal{U}_4, \mathcal{U}_5)\\
    \approx_\mathcal{C}
    SIM^{\mathcal{U}, t, k}_\mathcal{C}
    (x_\mathcal{C}, z, \mathcal{U}_1, \mathcal{U}_2, \mathcal{U}_3, \mathcal{U}_4, \mathcal{U}_5)      
    \end{aligned}
\end{equation*}
\begin{equation*}
    z=\left\{
    \begin{aligned}
        &\sum_{u\in\mathcal{U}_3\backslash\mathcal{C}^{x_u}},\ if |\mathcal{U}_3|\geq t,\\
        &\perp,\  otherwise
    \end{aligned}
    \right.
\end{equation*}
}
\end{theorem}

\revision{\textlabel{To}{R4:8:1} evaluate the privacy leakage from the aggregation of outputs, we employ the information gain metrics, in line with the methodologies used in~\cite{agrawal2001design, wang2012svd}. We define the privacy level as $\prod(Y|X):=2^{h(Y|X)}$, where $Y$ is the shared sum and $X$ denotes the original features that an adversary tries to deduce. Furthermore, the privacy leakage is defined as $\mathcal{P}(Y|X):=1-2^{h(Y|X)-h(Y)}$. Similar to~\cite{wang2012svd}, we take $\prod(Y|X)$ and $\mathcal{P}(Y|X)$ as privacy measures to quantify the privacy of our method. Table~\ref{tab:privacy_results} shows our privacy on the ``Cora'' dataset. Note that for Jester Data in~\cite{wang2012svd}, the highest privacy level is $6.5$, and our privacy level is significantly higher than theirs. It is because we only share the sum of the top k projection of the features, as opposed to the direct distortion of features by~\cite{wang2012svd}. For the privacy leakage,  our privacy leakage measure is substantially lower than 0.34, the least leakage reported by~\cite{wang2012svd} for Jester Data. }

\begin{table}[h]
\centering
\caption{Privacy Level and Leakage Across Different Numbers of Participants on ``Cora''}
\label{tab:privacy_results}
\begin{tabular}{ccc}
\hline
\textbf{Number of Participants} & \textbf{Privacy Level} & \textbf{Privacy Leakage} \\ \hline
2                               & 2438.21                & 0.09312                  \\
4                               & 2433.96                & 0.09311                  \\
6                               & 2424.31                & 0.09301                  \\
8                               & 2402.38                & 0.09311                  \\
10                              & 2372.59                & 0.09307                  \\
12                              & 2348.12                & 0.09305                  \\
14                              & 2321.91                & 0.09302                  \\
16                              & 2298.28                & 0.09299                  \\ \hline
\end{tabular}
\end{table}

\renewcommand*{\algorithmcfname}{Protocol}
\begin{algorithm}[tb]
\caption{Secure aggregation}
\label{alg:aggregation}
\small
\KwIn{each $\Cli_i$ inputs $x_i$}
\KwOut {$x =\sum\limits^l_{i=1}x_i$}
\BlankLine

\CommentSty{\bf Setup:} \hfill\CommentSty{once and for all rounds} \\
$\Cli_l$ initializes a large prime $p$, a group generator $g$, and a small modular $N$; ~sends them to other $\Cli$s

\For(\hfill \CommentSty{for each $\Cli$}){$i=1 \to l-1$}{
    $\Cli_i$: ~~$s_i\overset{\$}{\leftarrow}\mathbb{Z}_{p}$;~~
    $S_i\gets g^{s_i}\mod p$; ~~sends $S_i$ to $\Cli_l$
    
    \For(\hfill \CommentSty{for each $\Cli_j \neq \Cli_i$}) {$j = 1 \to l-1~\mathit{and}~j \neq i$} {
        $\Cli_l$ sends $S_i$ to $\Cli_j$
    }
}

\BlankLine
\CommentSty{\bf Aggregation ($k$th round)}

\For(\hfill \CommentSty{for each $\Cli$}){$i=1 \to l-1$}{
    
    \For(\hfill \CommentSty{for each $\Cli_j \neq \Cli_i$}) {$j = 1 \to l-1~\mathit{and}~j \neq i$} {
        $\Cli_i$: $S_{i, j} \gets S^{s_i}_j \mod p$;
        ~~~~~~~~~~~~~~~~~~~~~~$r_{i, j}\gets \alpha\cdot \mathsf{PRG}(k || S_{i, j})\mod N$ $(\mathit{if}~i < j,~\alpha = 1,~\mathit{else}~\alpha = -1)$\hfill \CommentSty{can be preprocessed}
    }
    $\Cli_i$: $\tilde{x}_i \gets (x_i + \sum\limits_{j\neq i}r_{i, j}) \mod N$; ~~sends $\tilde{x}_i$ to $\Cli_l$
}

$\Cli_l$: $x \gets (\sum\limits^{l-1}_{i=1}\tilde{x}_i + x_l) \mod N$


\end{algorithm}

\section{Evaluating kCAGC with split graph structures }\label{sec.split_graph}
\revision{
Given the situation where different participants possess their own graph structure, we study the case involving two participants to evaluate the effectiveness of our methods in this setting. However, since the absence of graph structure is anticipated to diminish the accuracy of the collaborative methods, we investigate how much shared graph data is required to make our method comparable to the centralized method.
We start with allocating $50\%$ of the complete graph structure to each participant, and gradually increase the proportion of the total graph distributed to them. Therefore, these two participants will gradually have more shared graph data.}

\begin{figure}
    \centering
    \includegraphics[width=.95\linewidth]{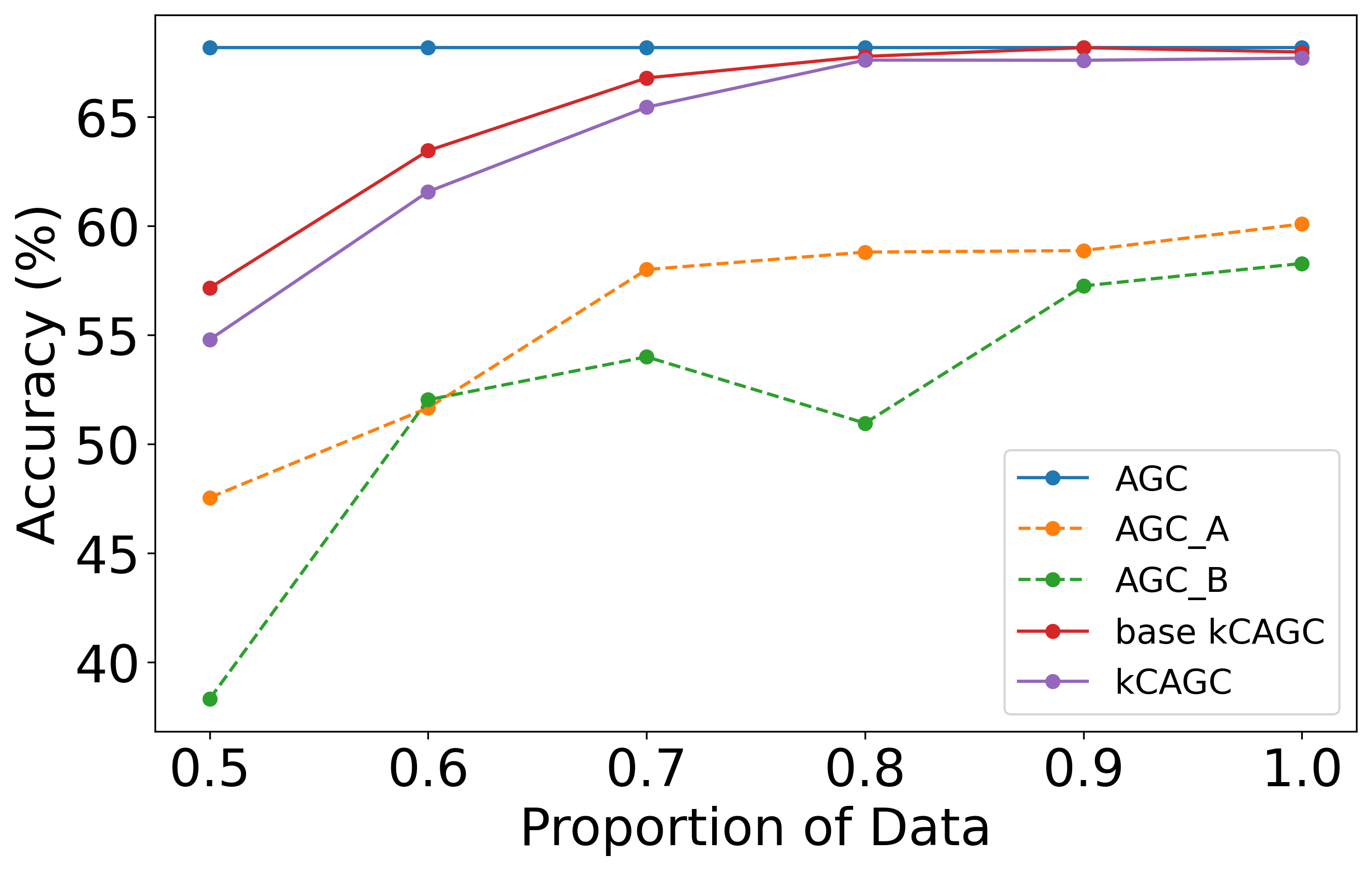}
    \caption{Accuracy of $\name$ while increasing the proportion of shared graph data for ``Cora'' Dataset}
    \label{fig:graphsplit}
\end{figure}
\revision{
Figure~\ref{fig:graphsplit} shows the results on the ``Cora'' Dataset. The experiment shows that both basic $\name$ and $\name$ outperform the models trained on isolated subgraphs. When each participant has more than $70\%$ of the complete graph structure, $\name$ can achieve similar accuracy compared to the centralized method. Using our methods can always achieve a better model compared to the models trained isolated ($AGC_A$ and $AGC_B$).}

\section{Additional Experiments}\label{appdix.add_exp}

\begin{figure}[ht]
\centering
\includegraphics[width=.95\linewidth]{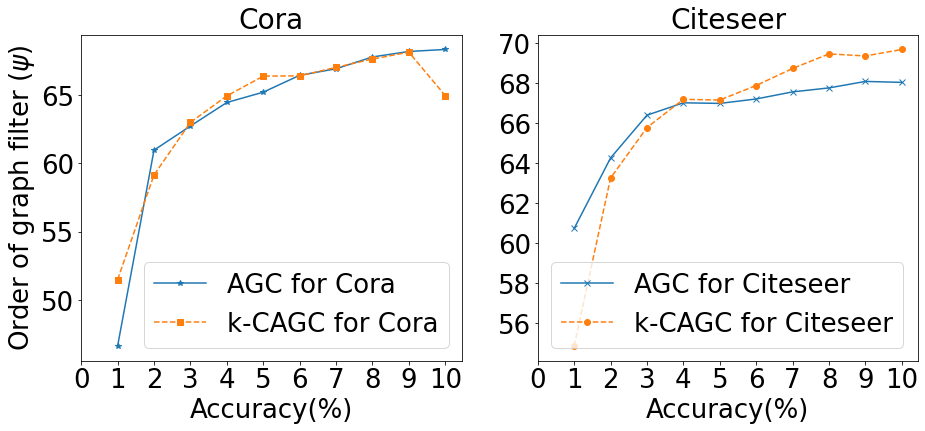}
\caption{Different order of graph filter ($\psi$) for $\name$ }
\label{fig.order}
\end{figure}
\subsection{Order of graph filter}
In our experiments, we set the number of participants to $L=2$ and the number of clusters to $\lc=k$. We compare the performance of our proposed method, denoted as $\name$, with that of the AGC method for different orders of graph filter $\psi$, while still applying the same graph filter for both methods. We conducted experiments on three datasets, including Cora, Citeseer, and Pubmed.

The results, shown in Figure~\ref{fig.order}, demonstrate that $\name$ can achieve similar performance to AGC, regardless of the order of the graph filter. This is because our method does not make any assumptions about the graph filter used. Instead, it adapts to the graph structure and the properties of the data itself. 

\subsection{Graph filter}

We further evaluated our method in comparison to centralized settings using different graph filters. Specifically, we selected the adaptive graph filter $p(\lambda)=(1-\frac{\lambda}{||L_s||})^\psi$ with $\psi = 5$ for both the "Cora" and "Citeseer" datasets. This filter is a nonnegative decreasing function for $\lambda\in[0,2]$ and was applied to the filtered node features as a baseline using Protocol~\ref{alg:centralized}.

We assessed the performance of our method using three widely adopted metrics: accuracy (ACC), normalized mutual information (NMI), and macro F1-score (F1). The results, presented in Table~\ref{tab:filter} for "Cora" and Table~\ref{tab:filter2} for "Citeseer", demonstrate that our method achieves similar performance to the baselines across all performance measures when using this new graph filter. Results in Table~\ref{tab:filter} and fig~\ref{fig.order} suggest that our approach is effective and can provide collaborative results for different graph filtering scenarios..

\begin{table*}[ht]
\caption{Summary of node classification results with a different garph filter for ``Cora'' dataset}
\label{tab:filter}
\centering
\begin{tabular}{@{}ccccccccccccc@{}}
\toprule
                              & \multicolumn{3}{c}{$L=2$}                                    & \multicolumn{3}{c}{$L=4$}                                    & \multicolumn{3}{c}{$L=8$}                                    & \multicolumn{3}{c}{$L=16$}              \\ \cmidrule(r){2-4}\cmidrule(r){5-7}\cmidrule(r){8-10}\cmidrule(r){11-13}
\multicolumn{1}{c|}{}         & ACC         & NMI         & \multicolumn{1}{c|}{F1}          & ACC         & NMI         & \multicolumn{1}{c|}{F1}          & ACC         & NMI         & \multicolumn{1}{c|}{F1}          & ACC         & NMI         & F1          \\ \midrule
\multicolumn{1}{c|}{$\lc=k$}  & $\mathbf{67.90}$      & $\mathbf{50.84}$       & \multicolumn{1}{c|}{$\mathbf{65.37}$}       & 66.30       & 52.43       & \multicolumn{1}{c|}{62.42}       & 66.45       & 52.00       & \multicolumn{1}{c|}{61.86}       & 65.94       & 50.59       & $\mathbf{59.76}$       \\
\multicolumn{1}{c|}{}         & $(\pm1.85)$ & $(\pm1.94)$ & \multicolumn{1}{c|}{$(\pm2.83)$} & $(\pm0.47)$ & $(\pm1.72)$ & \multicolumn{1}{c|}{$(\pm2.53)$} & $(\pm3.37)$ & $(\pm1.28)$ & \multicolumn{1}{c|}{$(\pm5.03)$} & $(\pm3.83)$ & $(\pm1.65)$ & $(\pm6.59)$ \\
\multicolumn{1}{c|}{$\lc=2k$} & 67.05       & 49.64       & \multicolumn{1}{c|}{62.93}       & $\mathbf{71.32}$       & $\mathbf{54.24}$       & \multicolumn{1}{c|}{$\mathbf{70.66}$}       & $\mathbf{68.21}$       & $\mathbf{53.77}$       & \multicolumn{1}{c|}{$\mathbf{62.86}$}       & $\mathbf{66.66}$       & $\mathbf{51.60}$       & 57.24       \\
\multicolumn{1}{c|}{}         & $(\pm1.93)$ & $(\pm1.10)$ & \multicolumn{1}{c|}{$(\pm3.72)$} & $(\pm1.42)$ & $(\pm0.59)$ & \multicolumn{1}{c|}{$(\pm1.89)$} & $(\pm3.16)$ & $(\pm1.67)$ & \multicolumn{1}{c|}{$(\pm7.01)$} & $(\pm2.85)$ & $(\pm1.42)$ & $(\pm3.96)$ \\
\multicolumn{1}{c|}{baseline} & 66.91       & 51.24       & \multicolumn{1}{c|}{63.94}       &             &             &                                  &             &             &                                  &             &             &             \\
\multicolumn{1}{c|}{}         & $(\pm0.04)$ & $(\pm0.04)$ & \multicolumn{1}{c|}{$(\pm0.08)$} &             &             &                                  &             &             &                                  &             &             &             \\ \bottomrule
\end{tabular}
\end{table*}

\begin{table*}[ht]
\caption{Summary of node classification results with a different garph filter for ``Citeseer'' dataset}
\label{tab:filter2}
\centering
\begin{tabular}{@{}ccccccccccccc@{}}
\toprule
                              & \multicolumn{3}{c}{$L=2$}                                    & \multicolumn{3}{c}{$L=4$}                                    & \multicolumn{3}{c}{$L=8$}                                    & \multicolumn{3}{c}{$L=16$}              \\ \midrule
\multicolumn{1}{c|}{}         & ACC         & NMI         & \multicolumn{1}{c|}{F1}          & ACC         & NMI         & \multicolumn{1}{c|}{F1}          & ACC         & NMI         & \multicolumn{1}{c|}{F1}          & ACC         & NMI         & F1          \\ \midrule
\multicolumn{1}{c|}{$\lc=k$}  & $\mathbf{70.24}$       & $\mathbf{44.47}$       & \multicolumn{1}{c|}{$\mathbf{65.30}$}       & 65.16       & 39.72       & \multicolumn{1}{c|}{60.39}       & $\mathbf{66.80}$       & 41.65       & \multicolumn{1}{c|}{$\mathbf{62.15}$}       & 69.10       & 43.32       & 63.61       \\
\multicolumn{1}{c|}{}         & $(\pm0.54)$ & $(\pm0.61)$ & \multicolumn{1}{c|}{$(\pm0.43)$} & $(\pm2.19)$ & $(\pm1.84)$ & \multicolumn{1}{c|}{$(\pm2.54)$} & $(\pm1.66)$ & $(\pm0.35)$ & \multicolumn{1}{c|}{$(\pm2.08)$} & $(\pm1.35)$ & $(\pm0.64)$ & $(\pm2.18)$ \\
\multicolumn{1}{c|}{$\lc=2k$} & 69.34       & 43.49       & \multicolumn{1}{c|}{64.61}       & $\mathbf{68.73}$       & $\mathbf{42.81}$       & \multicolumn{1}{c|}{$\mathbf{63.57}$}       & 66.49       & $\mathbf{41.75}$       & \multicolumn{1}{c|}{62.10}       & $\mathbf{71.04}$       & $\mathbf{45.08}$       & $\mathbf{64.23}$       \\
\multicolumn{1}{c|}{}         & $(\pm0.50)$ & $(\pm0.60)$ & \multicolumn{1}{c|}{$(\pm0.45)$} & $(\pm0.45)$ & $(\pm0.90)$ & \multicolumn{1}{c|}{$(\pm1.07)$} & $(\pm1.12)$ & $(\pm0.76)$ & \multicolumn{1}{c|}{$(\pm1.67)$} & $(\pm0.96)$ & $(\pm0.61)$ & $(\pm1.85)$ \\
\multicolumn{1}{c|}{baseline} & 67.52       & 41.55       & \multicolumn{1}{c|}{63.02}       &             &             &                                  &             &             &                                  &             &             &             \\
\multicolumn{1}{c|}{}         & $(\pm0.07)$ & $(\pm0.07)$ & \multicolumn{1}{c|}{$(\pm0.06)$} &             &             &                                  &             &             &                                  &             &             &             \\ \bottomrule
\end{tabular}
\end{table*}

\subsection{Effenciency in WAN}\label{appendix.E3}

We conduct our method on WAN of 400mbps bandwith with 100ms decay.
Table~\ref{tab:all_efficient_2} shows the training time in WAN. The results are similar to LAN. However, since we add 100ms decay in WAN, the shortest training time for the four datasets is about $10$ seconds. And for ``Wiki'' with $\lc=8K, L=16$, the training time is about $20$ minutes. However, when $\lc = k,2k$, the training time for all experiments will be within $1.5$ minutes.

\begin{table*}[ht]
\caption{Efficiency of k-CAGC in WAN. 
}
\label{tab:all_efficient_2}
\centering
\begin{tabular}{cccccccccc}
\hline
                               &      & \multicolumn{4}{c}{Cora}        & \multicolumn{4}{c}{Citeseer}      \\\cmidrule(r){3-6}  \cmidrule(r){7-10}\noalign{\smallskip} 
                               &      & $L=2$   & $L=4$    & $L=8$    & $L=16$   & $L=2$   & $L=4$    & $L=8$    & $L=16$    \\ \hline
\multirow{4}{*}{Training Time} & $\lc=k$  & 12.85 & 26.57 & 41.22  & 57.12  & 12.81 & 26.56  & 41.15  & 56.87   \\
                               & $\lc=2k$ & 13.06 & 26.79 & 42.46  & 59.16  & 13.00 & 26.78  & 42.34  & 58.79   \\
                               & $\lc=4k$ & 15.21 & 29.71 & 46.39  & 67.05  & 17.32 & 35.59  & 55.68  & 77.02   \\
                               & $\lc=8k$ & 24.23 & 43.29 & 71.34  & 117.51 & 20.33 & 37.84  & 60.92  & 96.46   \\ \hline
                               &      & \multicolumn{4}{c}{Pubmed}      & \multicolumn{4}{c}{Wiki}          \\ \cmidrule(r){3-6}  \cmidrule(r){7-10}\noalign{\smallskip} 
                               &      & $L=2$   & $L=4$    & $L=8$    & $L=16$   & $L=2$   & $L=4$    & $L=8$    & $L=16$    \\
                               \hline
\multirow{4}{*}{Training Time} & $\lc=k$  & 13.14 & 26.88 & 42.96  & 59.42  & 13.39 & 27.01  & 41.59  & 57.47   \\
                               & $\lc=2k$ & 13.65 & 27.62 & 43.38  & 59.85  & 18.16 & 33.59  & 53.05  & 78.46   \\
                               & $\lc=4k$ & 15.54 & 29.42 & 44.33  & 61.00  & 35.09 & 59.11  & 99.80  & 172.43  \\
                               & $\lc=8k$ & 19.00 & 34.39 & 52.83  & 73.77  & 92.88 & 281.89 & 741.55 & 1944.19 \\\hline
\end{tabular}

\end{table*}

%




\end{document}